\documentclass[lettersize,journal]{IEEEtran}
\usepackage{amsmath, amsfonts}
\usepackage{algorithmic}
\usepackage{algorithm}
\usepackage{colortbl}
\usepackage{xcolor}
\usepackage{array}
\usepackage[caption=false,font=normalsize,labelfont=sf,textfont=sf]{subfig}
\usepackage{textcomp}
\usepackage{stfloats}
\usepackage{url}
\usepackage{verbatim}
\usepackage{graphicx}
\usepackage{cite}
\usepackage[implicit=false]{hyperref}
\hypersetup{colorlinks=true, linkcolor=blue, urlcolor=blue}
\usepackage{threeparttable}
\usepackage{multirow}
\usepackage{arydshln}
\usepackage{booktabs}
\begin{document}

% -------- Title and Authors' Information: Begin -------- %
\title{Event Voxel Set Transformer for Spatiotemporal Representation Learning on Event Streams}

\author{Bochen Xie,~\IEEEmembership{Member,~IEEE}, Yongjian Deng,~\IEEEmembership{Member,~IEEE}, Zhanpeng Shao,~\IEEEmembership{Member,~IEEE},\\ Qingsong Xu,~\IEEEmembership{Senior Member,~IEEE}, and Youfu Li,~\IEEEmembership{Fellow,~IEEE}
        % <-this % stops a space
\thanks{This work was supported in part by the Research Grants Council of Hong Kong under Grant CityU11213420 and Grant CityU11206122; and in part by the National Natural Science Foundation of China under Grant 62173286, Grant 62203024, and Grant 61976191. \textit{(Bochen Xie and Yongjian Deng contributed equally to this work.)} \textit{(Corresponding author: Youfu Li.)}}
\thanks{Bochen Xie and Youfu Li are with the Department of Mechanical Engineering, City University of Hong Kong, Hong Kong SAR, China (e-mail: boxie4-c@my.cityu.edu.hk; meyfli@cityu.edu.hk).}
\thanks{Yongjian Deng is with the College of Computer Science, Beijing University of Technology, Beijing 100124, China (e-mail: yjdeng@bjut.edu.cn).}
\thanks{Zhanpeng Shao is with the College of Information Science and Engineering, Hunan Normal University, Changsha 410081, China (e-mail: zpshao@hunnu.edu.cn).}
\thanks{Qingsong Xu is with the Department of Electromechanical Engineering, University of Macau, Macao SAR, China (e-mail: qsxu@um.edu.mo).}
\thanks{Digital Object Identifier 10.1109/TCSVT.2024.3448615 }}% <-this % stops a space

\markboth{IEEE Transactions on Circuits and Systems for Video Technology
}{}

\IEEEpubid{\begin{minipage}{\textwidth}\ \centering
Copyright~\copyright~2024 IEEE. Personal use of this material is permitted. \\
However, permission to use this material for any other purposes must be obtained from the IEEE by sending an email to pubs-permissions@ieee.org.
\end{minipage}}
\maketitle

% \pagestyle{empty}  % no page number for the second and the later pages
% \thispagestyle{empty} % no page number for the first page
% --------Title and Authors' Information: End -------- %

% --------Abstract: Begin -------- %
\begin{abstract}
Event cameras are neuromorphic vision sensors that record a scene as sparse and asynchronous event streams. Most event-based methods project events into dense frames and process them using conventional vision models, resulting in high computational complexity. A recent trend is to develop point-based networks that achieve efficient event processing by learning sparse representations. However, existing works may lack robust local information aggregators and effective feature interaction operations, thus limiting their modeling capabilities. To this end, we propose an attention-aware model named Event Voxel Set Transformer (EVSTr) for efficient spatiotemporal representation learning on event streams. It first converts the event stream into voxel sets and then hierarchically aggregates voxel features to obtain robust representations. The core of EVSTr is an event voxel transformer encoder that consists of two well-designed components, including the Multi-Scale Neighbor Embedding Layer (MNEL) for local information aggregation and the Voxel Self-Attention Layer (VSAL) for global feature interaction. Enabling the network to incorporate a long-range temporal structure, we introduce a segment modeling strategy (S$^{2}$TM) to learn motion patterns from a sequence of segmented voxel sets. The proposed model is evaluated on two recognition tasks, including object classification and action recognition. To provide a convincing model evaluation, we present a new event-based action recognition dataset (NeuroHAR) recorded in challenging scenarios. Comprehensive experiments show that EVSTr achieves state-of-the-art performance while maintaining low model complexity.

\end{abstract}

% --------Abstract: End -------- %

% --------Introduction: Begin -------- %
\section{Introduction}\label{Sec_1_Introduction}

\begin{figure}[!t]
\centering
\includegraphics[width=3.45in]{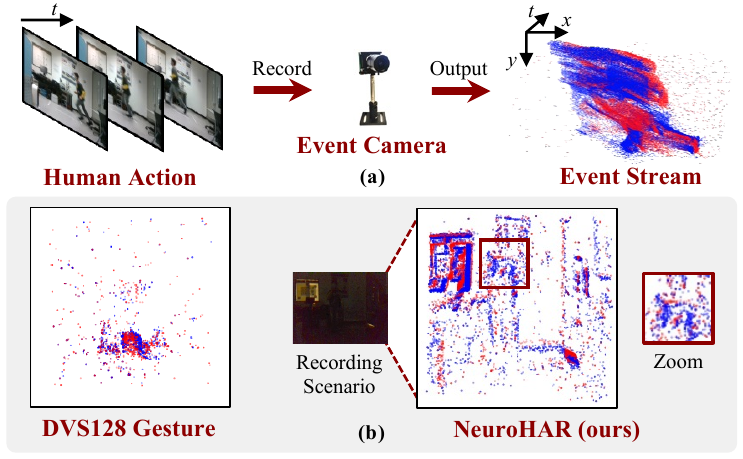}
\caption{(a) Visualization of event camera output. An event camera records the spatiotemporal information of the \textit{running} action as a stream of events, where red and blue dots denote positive and negative, respectively. (b) Comparison of samples from DVS128 Gesture \cite{amir2017dvsgesture} and NeuroHAR datasets. The sample is a human action \textit{hand clap}. After introducing handheld mobile recording, these samples in our proposed NeuroHAR contain both moving subject (zoom box) and background information. Besides, recording in the low-light illumination unlocks the high dynamic range advantage of event cameras.}
\label{Fig_Camera_and_Dataset}
\end{figure}

\IEEEPARstart{W}{ith} the rapid development of neuromorphic vision sensors \cite{lichtsteiner2008dvs128, posch2010atis, brandli2014davis240c}, event-based vision has broad and growing applications, including object classification \cite{gehrig2019est, cannici2020mlstm, messikommer2020asynet, deng2020amae, deng2021mvfnet, li2021evs, schaefer2022aegnn, deng2022evvgcnn, xie2022vmvgcn, wang2022dvsvit}, action recognition \cite{amir2017dvsgesture, wang2019eventcloud, bi2020rgcnn, liu2021motionsnn}, and optical flow estimation \cite{deng2021kd, liu2022edflow}. The event camera measures per-pixel brightness changes asynchronously and records visual information as a stream of events. The events are sparsely and unevenly distributed in the $x$-$y$-$t$ space, as shown in Fig. \ref{Fig_Camera_and_Dataset} (a). Each event consists of the spatial location, timestamp, and polarity of the brightness change. Compared to traditional cameras with a fixed frame rate, event cameras can offer significant advantages in the following aspects: low power consumption, high temporal resolution (in the order of $\mu$s), and high dynamic range ($>$ 120 dB) \cite{pami2022survey}. Thanks to these outstanding properties, the event camera unlocks great potential for recognition tasks in challenging visual scenarios, such as low-light illumination and high-speed movement. However, designing efficient learning algorithms for event-based recognition remains an open issue. According to the time duration of event streams, event-based recognition can be categorized into short-duration and long-duration recognition tasks, corresponding to object classification and action recognition, respectively. This paper focuses on developing an efficient learning model to flexibly recognize objects and human actions with event cameras.

\IEEEpubidadjcol
Existing event-based learning models can be summarized into two main categories: frame-based and point-based methods. Frame-based methods \cite{gehrig2019est, cannici2020mlstm, deng2020amae, deng2021mvfnet} first convert events into dense frame-based representations and then process them using learning models designed for images, such as convolutional neural networks (CNNs) and vision transformers. Combined with the models pre-trained on large-scale image datasets (e.g., ImageNet \cite{deng2009imagenet}), they achieve state-of-the-art performance on recognition tasks. However, the sparse-to-dense conversion sacrifices the sparse and asynchronous nature of event data, resulting in wasteful computation and high latency. To address this issue, point-based methods achieve a trade-off between computational cost and accuracy by fitting the nature of events. These approaches treat a single event or a set of regional events as the basic unit and represent event streams as spikes \cite{orchard2015hfirst, liu2021motionsnn}, point clouds \cite{sekikawa2019eventnet, wang2019eventcloud, chen2022ecsnet}, or graphs \cite{bi2020rgcnn, li2021evs, deng2022evvgcnn, xie2022vmvgcn, schaefer2022aegnn}. They achieve outstanding performance by learning spatiotemporal cues from sparse event representations. Most of them focus on developing local feature aggregators (e.g., graph convolution) to improve recognition performance. However, existing point-based methods have the following problems in feature representation. (\textit{i}) Their local feature aggregators may not fully exploit the relations (both positional and semantic) between neighbors and lack well-suited multi-scale modeling to capture robust local information. (\textit{ii}) The absence of long-range feature interaction limits their capabilities in global representation learning.

This work proposes a powerful yet lightweight model named Event Voxel Set Transformer (EVSTr) to solve the above problems. EVSTr can flexibly process both short- and long-duration event streams in a voxel-wise way for efficient recognition tasks, including object classification and action recognition. We adopt the event voxel set representation \cite{deng2022evvgcnn, xie2022vmvgcn} as input, which is robust to noise while maintaining the sparse structure. The core of EVSTr is the event voxel transformer encoder that hierarchically extracts spatiotemporal features from local to global through two novel designs, including Multi-Scale Neighbor Embedding Layer (MNEL) and Voxel Self-Attention Layer (VSAL). To tackle the first issue, MNEL jointly encodes positional and semantic relations between neighboring voxels into attention scores for multi-scale feature aggregation, thereby learning robust local representations. To solve the second problem, VSAL introduces absolute-relative positional encoding to assist vanilla self-attention operators in feature interaction between input elements with spatiotemporal structure, enabling better global modeling. We combine the encoder with a classification head to process a single voxel set converted from the event stream for object classification.

Furthermore, we extend this encoder with a Stream-Segment Temporal Modeling Module (S$^{2}$TM) to learn temporal dynamics for action recognition. Specifically, we split a long-duration event stream into a sequence of segmented voxel sets and extract spatiotemporal features per segment by the encoder. Then, a sequence of features is fed to S$^{2}$TM for aggregating long-range temporal dynamics. In this work, we also present a new event-based action recognition dataset named Neuromorphic Human Action Recognition (NeuroHAR) to alleviate the limitations of current benchmark datasets. Existing real-world action recognition datasets \cite{amir2017dvsgesture, liu2021motionsnn} are captured under relatively simple recording conditions (i.e., sufficient illumination and still cameras) that cannot reflect the actual application scenarios of event cameras, such as dark environment and dynamic background. To build a more challenging and practical dataset for convincing model evaluation, we collect NeuroHAR using an event camera and an RGB-D camera under challenging recording conditions, including low-light illumination and camera motion. An intuitive comparison of NeuroHAR and the widely used DVS128 Gesture dataset \cite{amir2017dvsgesture} is shown in Fig. \ref{Fig_Camera_and_Dataset} (b). NeuroHAR contains 1584 samples with 3 visual modalities covering 18 classes of daily actions. We evaluate the proposed model on four object classification datasets and five action recognition datasets to demonstrate its competitive accuracy and efficiency advantage.

The main contributions of this paper are summarized as follows:

\begin{itemize}
\item To fit the sparsity of events, we develop the EVSTr model that efficiently learns spatiotemporal representations on voxelized event streams for recognition tasks.
\item We introduce the event voxel transformer encoder with well-designed MNEL and VSAL layers to hierarchically extract spatiotemporal features from local to global. The segment modeling strategy (S$^{2}$TM) endows our network with a long-range temporal modeling capability.
\item We present a new event-based action recognition dataset (NeuroHAR) recorded under challenging conditions for providing a convincing model evaluation.
\item Extensive experiments on recognition tasks demonstrate that our method achieves state-of-the-art performance and has a significant advantage in model complexity.
\end{itemize}
% --------Introduction: End -------- %

% --------Related Work: Begin -------- %
\section{Related Work}

\subsection{Deep Learning on Event Streams}\label{Sec2_A_Event_Learning}

From the perspective of input representations, existing event-based learning models can be summarized into two mainstream strategies: frame-based and point-based methods.

\subsubsection{Frame-Based Methods} 
\begin{figure}[!t]
\centering
\includegraphics[width=3.3in]{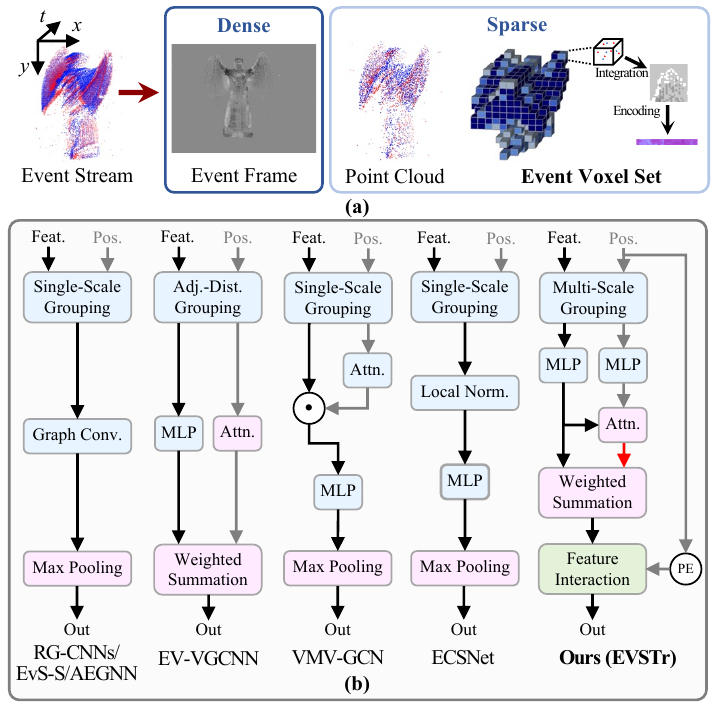}
\caption{(a) Visualization of the human action \textit{jumping jacks} with three representative event representations. (b) Comparison of different point-based architectures for object classification. Blue, pink and green blocks represent local feature encoding, aggregation function and global modeling, respectively. The competitors include RG-CNNs \cite{bi2020rgcnn}, EvS-S \cite{li2021evs}, AEGNN \cite{schaefer2022aegnn}, EV-VGCNN \cite{deng2022evvgcnn}, VMV-GCN \cite{xie2022vmvgcn}, and ECSNet \cite{chen2022ecsnet}. Our EVSTr has two important improvements: (\textit{i}) jointly modeling positional and semantic relations between neighbors to achieve multi-scale attentive aggregation; (\textit{ii}) leveraging self-attention to perform feature interaction in global modeling. $\odot$ denotes Hadamard product and PE is the abbreviation of positional encoding.}
\label{Fig_EvRep_Archs}
\end{figure}

This line of work first converts events within a time interval into dense frames and then processes them using off-the-shelf learning models tailored for images. Frame-based methods mainly focus on designing suitable representations to integrate the spatial semantics embedded in event streams. An example of the event frame is visualized in Fig. \ref{Fig_EvRep_Archs} (a). Some early works project events along the time-axis through handcrafted integration ways, such as cumulative quantity \cite{maqueda2018eventframe} and timestamp integration \cite{zhu2018voxelgrid}. Then, learnable representations \cite{gehrig2019est, cannici2020mlstm, uddin2022depth} replace the handcrafted integration kernels with trainable ones to achieve better generalization on downstream tasks. Besides, some works improve the robustness of representations from other perspectives, such as events-to-video reconstruction \cite{rebecq2019e2vid}, motion-agnostic filter \cite{deng2020amae}, and multi-view integration \cite{deng2021mvfnet}. Combined with pre-trained models, the frame-based family achieves state-of-the-art performance on many event-based tasks.

The conversion of dense frames sacrifices the sparse and asynchronous nature of events. Although frame-based methods outperform point-based ones in terms of accuracy, they usually require heavy-weight models to process dense input, leading to high computational complexity and latency. Some sparse frame-based solutions \cite{messikommer2020asynet, wang2022dvsvit} are proposed to tackle these limitations. They lag behind their dense frame-based counterparts in accuracy and cannot lead point-based methods in model complexity.

\subsubsection{Point-Based Methods} Unlike frame-based counterparts, point-based methods exploit the sparsity and asynchrony of event data and achieve a trade-off between accuracy and model complexity. They design sparse representations to convey the spatial and temporal information of event streams, as shown in Fig. \ref{Fig_EvRep_Archs} (a). A natural approach is to take events as input spikes and process them with spiking neural networks (SNNs) \cite{orchard2015hfirst, liu2021motionsnn} because the bio-inspired working mechanism of event cameras and SNNs is compatible \cite{pami2022survey}. However, training an SNN is more intractable than other learning architectures (e.g., CNNs) due to the lack of efficient optimization methods \cite{schaefer2022aegnn}. Inspired by 3D point cloud processing, some researchers adopt the event-specific variants of PointNet \cite{qi2017pointnet} or PointNet++ \cite{qi2017pointnet++} for several tasks, including semantic segmentation \cite{sekikawa2019eventnet} and action recognition \cite{wang2019eventcloud}. However, these network architectures have limited capabilities in capturing local features because they directly follow the point cloud processing manner, which does not fit the spatiotemporal structure of events well. ECSNet \cite{chen2022ecsnet} adopts the 2D-1T event cloud sequence as an intermediate representation, enhancing performance by better aligning with the nature of event data.

Recent graph-based approaches \cite{bi2020rgcnn, li2021evs, schaefer2022aegnn} construct point-wise graphs on downsampled event streams and exploit graph neural networks (GNNs) to extract event features. Considering that the point-wise relationship of events is susceptible to noise signal \cite{xu2023denoising}, voxel-wise GNNs \cite{deng2022evvgcnn, xie2022vmvgcn} first convert an event stream into a voxel set and then aggregate vertices' information for feature representation. The voxel-wise representations keep more local semantics than the point-wise ones, gaining the robustness to noise. As stated in Section \ref{Sec_1_Introduction}, existing point-based methods still have some limitations that need to be improved, mainly the defective design of local feature aggregators and the lack of feature interaction. As shown in Fig. \ref{Fig_EvRep_Archs} (b), we compare our EVSTr with some representative architectures to better demonstrate the improvement. Instead of aggregating neighbor features relying only on positional relation, the proposed MNEL layer comprehensively considers positional and semantic correlations to adaptively aggregate over multi-scale neighborhood spaces, thus capturing robust local features. For better global modeling, our VSAL layers leverage self-attention with a novel positional encoding approach to perform feature interaction. Additional segment modeling extend our network's long-range temporal learning capability for human action recognition.

\subsection{Transformers in Vision}\label{Sec2_B_Transformers_in_CV}

The great success of transformer architectures \cite{vaswani2017transformer} in natural language processing attracts increasing interest among computer vision researchers. The core component of transformers is the self-attention mechanism that computes semantic correlations between input elements to model long-range dependencies. Researchers have recently applied transformers to many vision tasks, such as image recognition \cite{dosovitskiy2021vit, liu2021swin}, video classification \cite{bertasius2021timesformer, arnab2021vivit, neimark2021vtn}, point cloud processing \cite{guo2021pct, zhao2021pt}, and event stream processing \cite{wang2022dvsvit, tian2022evtflownet}. Previous event-based self-attention networks take event frames as input and process them with vision transformers designed for images. Our interest is applying self-attention to sparse event representations for efficient event processing. Inspired by the point cloud transformer \cite{guo2021pct}, we follow the encoder design strategy from local to global modeling and propose a novel method with event-specific designs, such as multi-scale attentive aggregation for voxel-wise neighbor embedding and absolute-relative positional encoding for assisting self-attention in operating on spatiotemporal data. Besides, we learn long-range temporal dynamics by a self-attention based segment modeling strategy for action recognition. Eventually, we develop EVSTr with a well-designed transformer encoder and segment modeling strategy, which can efficiently learn spatiotemporal representations for event-based recognition.
% --------Related Work: End -------- %

% --------Methodology: Begin -------- %
\begin{figure*}[htbp]
\centering
\includegraphics[width=6.9in]{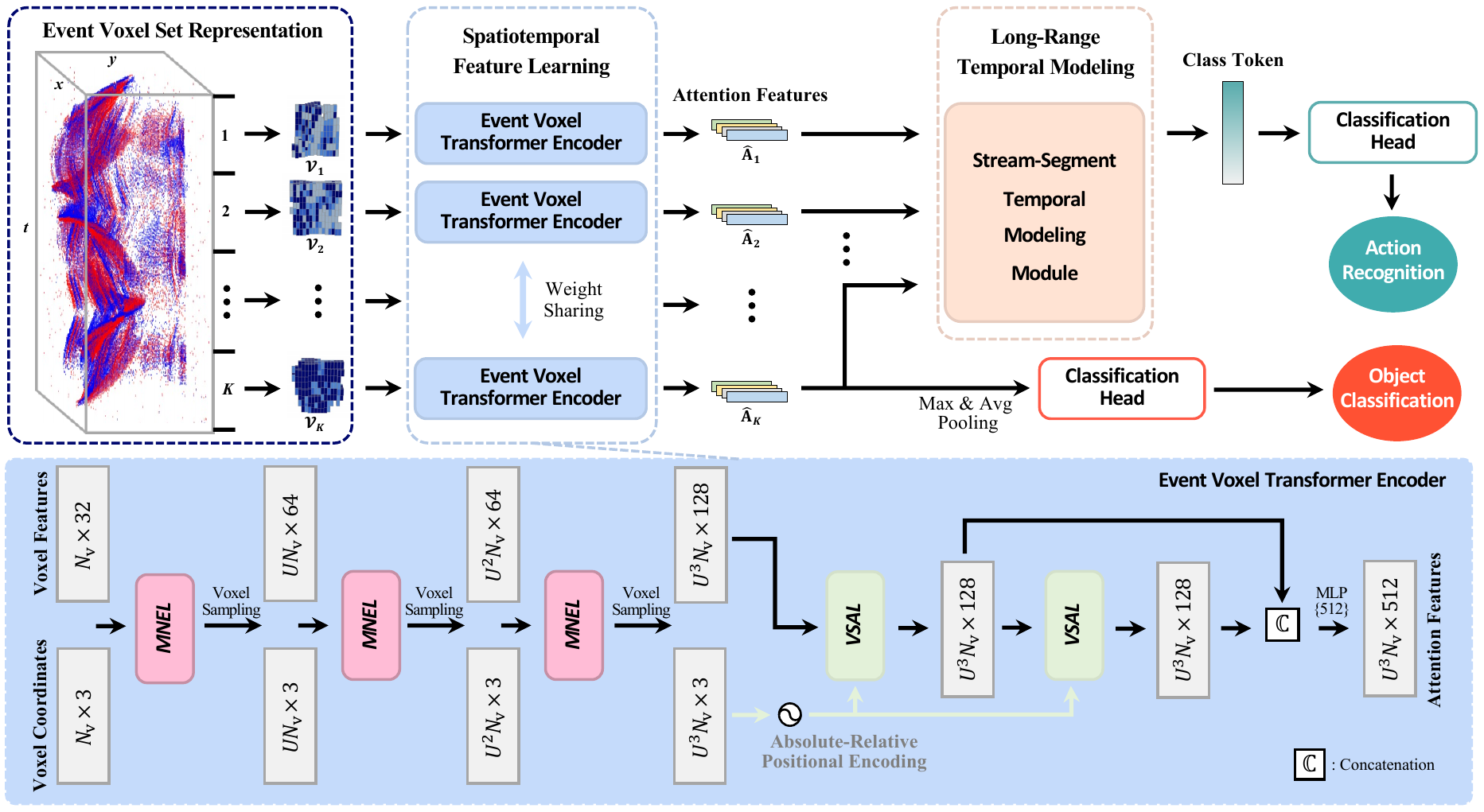}
\caption{The pipeline of Event Voxel Set Transformer (EVSTr) for object classification and action recognition. We spilt an event stream into $K$ segments of equal duration ($K = 1$ for object classification) and convert them as event voxel sets $\{\mathcal{V}_{1}, \mathcal{V}_{2}, ..., \mathcal{V}_{K}\}$. The event voxel transformer encoder propagates each voxel set to extract spatiotemporal features $\{\hat{\mathbf{A}}_{1}, \hat{\mathbf{A}}_{2}, ..., \hat{\mathbf{A}}_{K}\}$. For object classification, we feed the encoder output into a classification head for predicting classes. For action recognition, we first model long-range temporal dynamics over multiple voxel sets by a Stream-Segment Temporal Modeling Module and then transform the class token into categories. $N_{\rm v}$ represents the total number of input voxels, and $U$ denotes the downsampling rate of voxel sampling.}
\label{Fig_Model}
\end{figure*}

\section{Methodology}

An overview of the proposed EVSTr model is illustrated in Fig. \ref{Fig_Model}. EVSTr has two working modes corresponding to short-duration and long-duration recognition with event cameras. It comprises three stages: event representation, spatiotemporal feature learning, and long-range temporal modeling. For object classification, we convert an event stream into a single event voxel set, whereas for action recognition with long-range temporal context, we construct a sequence of segmented voxel sets over time duration. The event voxel transformer encoder processes each voxel set and extracts spatiotemporal features through learning inter-voxel relations. When recognizing human actions on long-duration streams, our EVSTr model combines parallel encoders with a segment modeling strategy to learn temporal dependencies from the input sequence of voxel sets. For short-duration object classification, EVSTr removes long-range temporal modeling and uses an encoder to extract appearance information from a single voxel set.

Section \ref{Sec3_A_Representation} introduces the event camera's working principle and event voxel set representation. The core components of the event voxel transformer encoder (MNEL and VSAL) and the process of spatiotemporal feature learning are detailed in Section \ref{Sec3_B_Encoder}. We introduce the long-range temporal modeling for action recognition in Section \ref{Sec3_C_S2TM}. Finally, we detail the network architectures of our model applied to object classification and action recognition in Section \ref{Sec3_D_Architecture}.

\subsection{Event Voxel Set Representation}\label{Sec3_A_Representation}

The event camera triggers an event $\mathbf{e}_{k} = (x_{k}, y_{k}, t_{k}, p_{k})$ when the logarithmic brightness $L$ at pixel $(x_{k}, y_{k})$ and time $t_{k}$ changes by a certain threshold $C$ ($C > 0$) \cite{pami2022survey}:
\begin{equation}
L(x_{k}, y_{k}, t_{k}) - L(x_{k}, y_{k}, t_{k} - \Delta t_{k}) = p_{k}C,
\end{equation}
where $\Delta t_{k}$ is the time interval between the last and current events at $(x_{k}, y_{k})$, and the range of $(x_{k}, y_{k})$ belongs to the camera's spatial resolution $H \times W$, and $p_{k} \in \{ +1, -1 \}$ is the polarity of brightness change (+1 for increasing and -1 for decreasing). An event can be treated as a space-time point at the coordinate $(x_{k}, y_{k}, t_{k})$ with a polarity attribute $p_{k}$.

As shown in Fig. \ref{Fig_EvRep_Archs} (a), we adopt the voxel-wise representation \cite{deng2022evvgcnn} to convert an event stream $\mathcal{E} = \{\mathbf{e}_{k}\}_{M}$ into a voxel set $\mathcal{V} = \{(\mathbf{f}_{i}, \mathbf{c}_{i})\}_{N_{\rm v}}$, where $M$ and $N_{\rm v}$ are the number of events and voxels. $\mathbf{f}_{i} \in \mathbb{R}^{D_{\rm f}}$ and $\mathbf{c}_{i} = (x_{i}^{\rm c}, y_{i}^{\rm c}, t_{i}^{\rm c}) \in \mathbb{R}^{3}$ are the voxel's feature and spatiotemporal coordinate. Specifically, we normalize the temporal size of event stream $\{\mathbf{e}_{k}\}_{M}$ with a compensation coefficient $T$ by $t_{k}^{*} = T (t_{k} - t_{1}) / (t_{M} - t_{1})$. After normalization, the event stream encompasses 3D space with the size of $H \times W \times T$. We define the spatiotemporal size of voxels as $H_{\rm v} \times W_{\rm v} \times T_{\rm v}$ to subdivide the event stream into a 3D voxel grid and select $N_{\rm v}$ non-empty voxels via motion-sensitive sampling \cite{xie2022vmvgcn}. We regard the location of each voxel within the 3D grid as a spatiotemporal coordinate. Denote $\{(x_{j}^{\rm v}, y_{j}^{\rm v}, t_{j}^{\rm v}, p_{j}^{\rm v})\}_{M_{\rm v}}$ as the internal events of the $i$th voxel, where $M_{\rm v}$ is the number of events and $i \in \{1, 2, ..., N_{\rm v}\}$. We can represent the voxel information by integrating internal events into a 2D patch $\mathbf{F}_{i} \in \mathbb{R}^{H_{\rm v} \times W_{\rm v}}$ along the time-axis. The value of each pixel $(x, y)$ in the patch is calculated by accumulating the timestamps and polarity of events \cite{deng2022evvgcnn}:
\begin{equation}
\begin{gathered}
\mathbf{F}_{i}(x, y) = \sum_{j}^{M_{\rm v}} \delta (x - x_{j}^{\rm v}, y - y_{j}^{\rm v}) p_{j}^{\rm v} t_{j}^{\rm v}, \\
\delta(a, b) = \begin{cases}
1,  & \text{if a = 0 and b = 0} \\
0, & \text{otherwise}
\end{cases},
\end{gathered}
\end{equation}
where $\delta$ is the Kronecker delta function \cite{maqueda2018eventframe, wang2022dvsvit} used to collect events at the same spatial location and the time interval of integration is $T_{\rm v}$ (i.e., voxel grid's temporal size). Then, we flatten $\mathbf{F}_{i}$ into a 1D vector in $\mathbb{R}^{H_{\rm v} W_{\rm v}}$, and use a multi-layer perceptron (MLP) to encode it into voxel features $\mathbf{f}_{i} \in \mathbb{R}^{D_{\rm f}}$:
\begin{equation}
\mathbf{f}_{i} = \mathtt{MLP} (\Gamma (\mathbf{F}_{i})),
\end{equation}
where $\Gamma$ is the flattening operation and $\mathtt{MLP}$ consists of a linear layer, a batch normalization (BN) and a ReLU function. Eventually, we obtain the event voxel set representation $\mathcal{V} = \{(\mathbf{f}_{i}, \mathbf{c}_{i})\}_{N_{\rm v}}$ not only retains local semantics inside voxels but also keeps the sparse structure of event data.

When processing short-duration event streams, it is reasonable to directly convert the entire event stream into a single voxel set. For a long-duration event stream, we spilt it into multiple segments of equal duration and convert them as voxel-wise representations individually to preserve longer temporal dynamics, as detailed in Section \ref{Sec3_C_S2TM}.

\subsection{Spatiotemporal Feature Learning}\label{Sec3_B_Encoder}

The event voxel set is fed into the event voxel transformer encoder, where our model learns spatiotemporal features from input elements, as illustrated in Fig. \ref{Fig_Model}. The encoder consists of two carefully designed components: Multi-Scale Neighbor Embedding Layer (MNEL) for local information aggregation and Voxel Self-Attention Layer (VSAL) for global feature interaction. In the following, we successively detail their design strategies and how to build the encoder with them.

\begin{figure*}[htbp]
\centering
\includegraphics[width=7in]{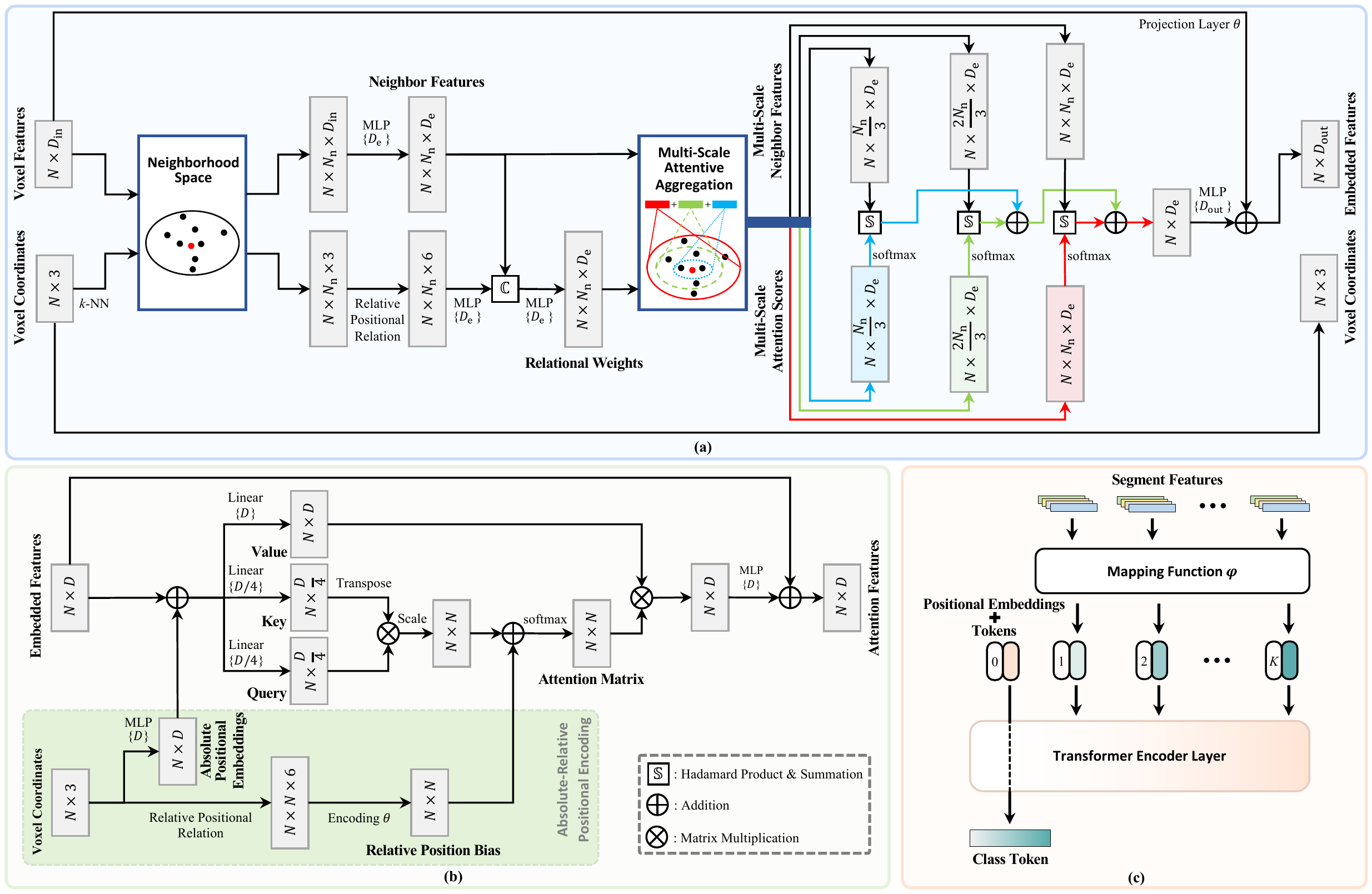}
\caption{The architecture of three components in the EVSTr model. (a) Multi-Scale Neighbor Embedding Layer (MNEL). It attentively aggregates multi-scale neighbor features into a local representation for each voxel. (b) Voxel Self-Attention Layer (VSAL). It performs inter-voxel feature interactions to enhance global representation. We introduce the absolute-relative positional encoding for assisting the self-attention operator in fitting with the spatiotemporal structure. (c) Stream-Segment Temporal Modeling Module (S$^{2}$TM). It is designed for action recognition by learning the long-range temporal dynamics from a sequence of segmented features.}
\label{Fig_Design}
\end{figure*}

\subsubsection{Multi-Scale Neighbor Embedding Layer} As illustrated in Fig. \ref{Fig_Design} (a), MNEL jointly encodes positional and semantic relations into attention scores for adaptive feature aggregation over multi-scale neighborhood spaces.

Given an event voxel set $\mathcal{V} = \{(\mathbf{f}_{i}, \mathbf{c}_{i})\}_{N}$, MNEL learns a local representation for each voxel by embedding multi-scale neighbor features, where $N$ is the number of input voxels for encoding layers. For the $i$th voxel, we gather its $N_{\rm n}$ neighboring voxels by the $k$-nearest neighbours ($k$-NN) algorithm based on the Euclidean distance of coordinates. This $k$-NN set includes a self-loop, which means the central voxel also links itself as a neighbor. In this neighborhood space, $\mathbf{f}_{ij} \in \mathbb{R}^{D_{\rm in}}$ and $\mathbf{c}_{ij} \in \mathbb{R}^{3}$ indicate the feature and spatiotemporal coordinate of the central voxel's $j$th neighbor, where $j \in \{1, 2, ..., N_{\rm n}\}$. We define the relative positional relation of a neighboring pair as $\mathbf{r}_{ij} = \mathbb{C}(\mathbf{c}_{i}, (\mathbf{c}_{i} - \mathbf{c}_{ij})) \in \mathbb{R}^{6}
$, where $\mathbb{C}$ is the concatenation operation. Then we encode the feature and relative positional relation vector of the $j$th neighbor using two MLP layers individually:
\begin{equation}
\tilde{\mathbf{f}}_{ij} = \mathtt{MLP} (\mathbf{f}_{ij}) \in \mathbb{R}^{D_{\rm e}},\ \tilde{\mathbf{r}}_{ij} = \mathtt{MLP} (\mathbf{r}_{ij}) \in \mathbb{R}^{D_{\rm e}}.
\end{equation}
As shown in Fig. \ref{Fig_EvRep_Archs} (b), previous works aggregate local features directly \cite{qi2017pointnet++, bi2020rgcnn, li2021evs, schaefer2022aegnn, xie2022vmvgcn} or only consider positional relations \cite{deng2022evvgcnn} as the correlation between neighboring pairs. These methods may be biased because the semantic relation is also critical in assessing the importance of each neighbor.

To comprehensively model the inter-voxel relationship, we fuse positional and semantic cues to generate the relational weights $\hat{\mathbf{r}}_{ij} \in \mathbb{R}^{D_{\rm e}}$ of this neighboring pair by
\begin{equation}
\hat{\mathbf{r}}_{ij} = \mathtt{MLP} (\mathbb{C} (\tilde{\mathbf{f}}_{ij}, \tilde{\mathbf{r}}_{ij})).
\end{equation}
For the central voxel, we obtain a $k$-NN set $\{(\tilde{\mathbf{f}}_{ij}, \hat{\mathbf{r}}_{ij})\}_{j = 1}^{{N_{\rm n}}}$ of neighbor features with corresponding relational weights. Inspired by PointNet++ \cite{qi2017pointnet++}, we propose a novel multi-scale attentive aggregation strategy to enhance local feature extraction. Based on the distance of neighbors to the central voxel, we divide the $k$-NN set into $S$ subspaces with increasing spatiotemporal scales. Our strategy aims to embed the features from multi-scale neighborhood subspaces into a local representation of the central voxel. In each subspace, we transform the relational weights of neighboring pairs into attention scores through the softmax function. After that, there are $S$ groups of attention scores $ {\textstyle \bigcup_{k=1}^{S}} \{\hat{\mathbf{r}}_{ijk}\}_{j=1}^{kN_{\rm n} / S}$ used to re-weight per-subspace neighbor features for aggregation, where $\hat{\mathbf{r}}_{ijk} \in \mathbb{R}^{D_{\rm e}}$ denotes the attention score of a neighboring pair in the $k$th group. We combine the features of different scales to achieve multi-scale embedding. This process can be formalized as 
\begin{equation}
\hat{\mathbf{f}}_{i} = \mathtt{MLP} (\sum_{k = 1}^{S} \sum_{j = 1}^{k N_{\rm n} / S} \tilde{\mathbf{f}}_{ij} \odot \hat{\mathbf{r}}_{ijk}),
\end{equation}
where $\hat{\mathbf{f}}_{i} \in \mathbb{R}^{D_{\rm out}}$ is a local representation of the $i$th voxel and $\odot$ denotes Hadamard product. Finally, we use a shortcut connection to fuse the input and embedded feature as output $\mathbf{l}_{i}$ by
\begin{equation}
\mathbf{l}_{i} = \hat{\mathbf{f}}_{i} + \theta (\mathbf{f}_{i}),
\end{equation}
where $\theta$ is a projection layer to guarantee that $\hat{\mathbf{f}}_{i}$ and $\mathbf{f}_{i}$ have the same dimension. When $D_{\rm out}$ is equal to $D_{\rm in}$, it is an identity projection, otherwise it is an MLP to match the dimension. For efficiency, the number $S$ of subspaces is 3, and we set $D_{\rm e}$ as $D_{\rm in}/2$.

Finally, a new set of event voxels $\mathcal{N} = \{(\mathbf{l}_{i}, \mathbf{c}_{i})\}_{N}$ is updated by MNEL, which attentively encodes the local representation of each voxel via embedding multi-scale neighbor features. After each MNEL layer, we utilize random sampling \cite{deng2022evvgcnn}, a simple yet efficient approach, to decrease voxel density and progressively enlarge the receptive field of feature aggregation. The downsampling rate of voxel sampling is a constant $U$. In the event voxel transformer encoder, we stack multiple MNEL layers and voxel sampling operations to hierarchically learn local representations and reduce the scale of voxel sets.

\subsubsection{Voxel Self-Attention Layer} After neighbor embedding with several MNEL layers, the encoder acquires the capability of local feature representation. We introduce the VSAL layer to handle feature interaction between voxels via self-attention, as shown in Fig. \ref{Fig_Design} (b).

Formally, given the input $\mathcal{N} = \{(\mathbf{l}_{i}, \mathbf{c}_{i})\}_{N}$, VSAL calculates semantic aﬃnities between voxels and learn a global representation via feature interaction. Vanilla self-attention cannot capture the order of input sequence, so incorporating position information is especially important for transformer networks. We introduce a trainable absolute-relative positional encoding approach that explicitly injects position information into the self-attention operator to capture the spatiotemporal order of voxels. Denoting that the input features are $\mathbf{L} = \{\mathbf{l}_{i}\}_{N} \in \mathbb{R}^{N \times D}$ and spatiotemporal coordinates are $\mathbf{C} = \{\mathbf{c}_{i}\}_{N} \in \mathbb{R}^{N \times 3}$ in the set $\mathcal{N}$, we first add the absolute positional embeddings $\mathbf{P} \in \mathbb{R}^{N \times D}$ to $\mathbf{L}$ and then map them into query ($\mathbf{Q}$), key ($\mathbf{K}$), and value ($\mathbf{V}$) matrices by linear transformations:
\begin{equation}
\begin{gathered}
\mathbf{P} = \mathtt{MLP} (\mathbf{C}), \\
\{\mathbf{Q}, \mathbf{K}, \mathbf{V}\} = (\mathbf{L} + \mathbf{P}) \{\mathbf{W}_{\rm q}, \mathbf{W}_{\rm k}, \mathbf{W}_{\rm v}\},
\end{gathered}
\end{equation}
where $\mathbf{W}_{\rm q}, \mathbf{W}_{\rm k} \in \mathbb{R}^{D \times (D/4)}$ and $\mathbf{W}_{\rm v} \in \mathbb{R}^{D \times D}$ are the learnable weights of linear layers. Inspired by \cite{liu2021swin}, we include a relative position bias $\mathbf{B} \in \mathbb{R}^{N \times N}$ to the scaled dot-product attention \cite{vaswani2017transformer} in computing similarity. Specifically, we calculate the relative positional relation between two voxels and design an encoding function to generate the bias matrix:
\begin{equation}
\mathbf{B} (i, j) = \phi (\mathbb{C} (\mathbf{c}_{i}, (\mathbf{c}_{i} - \mathbf{c}_{j}))),
\end{equation}
where $i = \{1, 2, ..., N\}$ and $j = \{1, 2, ..., N\}$ denote the row and column index of $\mathbf{B}$. Here, $\phi$ is a positional encoding function consisting of two linear layers separated by a BN layer and a ReLU function, which first projects the vector of relative positional relation into the latent space and then squeezes it to a scalar. Eventually, we employ the scaled dot-product attention equipped with the relative position bias to model input representations based on the interactions between all elements:
\begin{equation}
\mathbf{A} = \mathtt{softmax} (\frac{\mathbf{Q} \mathbf{K}^{\rm T}}{\sqrt{D} } + \mathbf{B}) \mathbf{V},
\end{equation}
where $\mathbf{A} \in \mathbb{R}^{N \times D}$ is a set of self-attention features retaining long-range interaction. We take an MLP to project $\mathbf{A}$ and add input features $\mathbf{L}$ to it as the output features:
\begin{equation}
\tilde{\mathbf{A}} = \mathtt{MLP} (\mathbf{A}) + \mathbf{L}.
\end{equation}

The VSAL layer achieves the goal of $\mathcal{N} \rightarrow \tilde{\mathbf{A}}$ transformation through adaptive feature interaction, thus obtaining superior global representation performance.

\subsubsection{Encoder} The architecture of event voxel transformer encoder is shown in Fig. \ref{Fig_Model}. The encoder contains two stages: neighbor embedding and global feature modeling. In the first stage, we stack three MNEL layers and voxel sampling operations to hierarchically incorporate local information into the embedded features. Then, the embedded features with corresponding coordinates are fed into two stacked VSAL layers to model global semantics. The number $N$ of input voxels for these five layers is $\{N_{\rm v}, UN_{\rm v}, U^{2}N_{\rm v}, U^{3}N_{\rm v}, U^{3}N_{\rm v}\}$. We concatenate the outputs of two VSAL layers and use an MLP to encode them as the final attention features $\hat{\mathbf{A}} \in \mathbb{R}^{U^{3}N_{\rm v} \times D}$. Eventually, the encoder can extract semantic information and short-term temporal dynamics by efficient spatiotemporal feature learning.

\subsection{Long-Range Temporal Modeling}\label{Sec3_C_S2TM}

In contrast to object classification, which mainly focuses on appearance information, recognizing human actions requires the model with a long-range temporal modeling capability. When the duration becomes long, how to model complex motion patterns on event streams is still challenging. Inspired by the advances in video processing methods \cite{wang2018tsn, arnab2021vivit, neimark2021vtn}, we introduce a segment modeling strategy to incorporate a long-range temporal structure for event-based action recognition, as shown in Fig. \ref{Fig_Design} (c).

Given a long-duration event stream $\mathcal{E}$, we spilt it into $K$ segments $\{\mathcal{E}_{1}, \mathcal{E}_{2}, ..., \mathcal{E}_{K}\}$ of equal duration. Each stream segment $\mathcal{E}_{K}$ is first converted into an event voxel set $\mathcal{V}_{K}$ and then fed into the weight-shared event voxel transformer encoder to learn short-term temporal dynamics $\hat{\mathbf{A}}_{K}$. We propose the Stream-Segment Temporal Modeling Module (S$^{2}$TM) that takes a sequence $\{\hat{\mathbf{A}}_{1}, \hat{\mathbf{A}}_{2}, ..., \hat{\mathbf{A}}_{K}\}$ as input to learn the inter-segment dependencies. S$^{2}$TM maps each segment's feature into a 1D token and utilizes self-attention to aggregate temporal dynamics. In specific, $\hat{\mathbf{A}}_{K}$ is integrated into a token through a mapping function $\varphi: \mathbb{R}^{U^{3}N_{\rm v} \times D} \rightarrow \mathbb{R}^{D}$. This function concatenates the outputs of max and average pooling on $\hat{\mathbf{A}}_{K}$ and then encodes the feature vector using an MLP. We prepend a learnable class token $\mathbf{s}_{\rm cls} \in \mathbb{R}^{D}$ to the sequence and add trainable positional embeddings $\mathbf{P}_{\rm t} \in \mathbb{R}^{(K + 1) \times D}$ \cite{dosovitskiy2021vit} to these tokens for retaining temporal information. Finally, a temporal modeling function processes the sequence to aggregate long-range temporal context. The workflow of S$^{2}$TM is summarized as follows:
\begin{equation}
\{\hat{\mathbf{s}}_{\rm cls}, \hat{\mathbf{s}}_{1}, ..., \hat{\mathbf{s}}_{K}\} = h (\{\mathbf{s}_{\rm cls}, \varphi (\hat{\mathbf{A}}_{1}), ..., \varphi (\hat{\mathbf{A}}_{K})\} + \mathbf{P}_{\rm t}),
\end{equation}
where $\{\hat{\mathbf{s}}_{\rm cls}, \hat{\mathbf{s}}_{1}, ..., \hat{\mathbf{s}}_{K}\}$ is a sequence of output tokens, and the temporal modeling function $h$ is a transformer encoder layer \cite{dosovitskiy2021vit} consisting of multi-head attention layer and feed-forward network. After feature propagation, we use the class token $\hat{\mathbf{s}}_{\rm cls} \in \mathbb{R}^{D}$ embedded with global temporal context as the final representation for prediction.

\subsection{Network Architecture}\label{Sec3_D_Architecture}

As visualized in Fig. \ref{Fig_Model}, EVSTr can be applied to both object classification and action recognition tasks. We detail how to construct EVSTr using the proposed encoder and long-range temporal modeling strategy.

\subsubsection{Object Classification} The key challenge of event-based object classification is extracting discriminative appearance information from short-duration event streams. We convert the entire stream into a single voxel set as input of the classification network. The input is fed into the event voxel transformer encoder to obtain semantic features via spatiotemporal feature learning. We use max and average pooling operations to process the features and concatenate them as a 1D vector. Finally, we use a classification head consisting of three fully connected (FC) layers to transform the global feature vector for categorization. Each FC layer is followed by a BN layer and a ReLU function, except the last one.

\subsubsection{Action Recognition} Unlike object classification, which depends on appearance information, a core challenge of event-based action recognition is learning temporal dynamics. To retain the long-range temporal structure, we split an event stream into multiple segments for voxel set representations separately and then model the inter-segment temporal dependencies. We utilize the event voxel transformer encoder to extract short-term temporal dynamics from each segment in a weight-shared manner. Then we feed the sequence of segmented features into S$^{2}$TM to aggregate long-range temporal context. At last, a single FC layer is used to process the class token into a final predicted category.
% -------- Methodology: End -------- %

% --------Experiments: Begin -------- %
\section{Experiments}

In this section, we evaluate the proposed EVSTr model on two event-based recognition tasks, including object classification in Section \ref{Sec4_A_Object} and action recognition in \ref{Sec4_B_Action}. Object classification effectively evaluates the algorithm's ability to extract spatiotemporal features from short-duration event streams, while action recognition on long-duration streams assesses the method's capability to model long-range temporal dependencies. To provide a more practical and convincing model evaluation, we present a new event-based action recognition dataset recorded in challenging visual scenarios. Besides, a detailed ablation study is provided in Section \ref{Sec4_C_Ablation} to analyze the effectiveness of different designs.

\subsection{Object Classification}\label{Sec4_A_Object}

\subsubsection{Dataset} We validate our method on four representative event-based object classification datasets: N-Caltech101 \cite{orchard2015ncal}, CIFAR10-DVS \cite{li2017cifar10dvs}, N-Cars \cite{sironi2018hats} and ASL-DVS \cite{bi2020rgcnn}. N-Caltech101 records the RGB images using a moving event camera. CIFAR10-DVS records the moving RGB images displayed on a monitor via a still event camera. Instead, N-Cars and ASL-DVS uses event cameras to record real-world objects. We train our model on each training set separately and evaluate its performance on the testing sets. For N-Caltech101, CIFAR10-DVS, and ASL-DVS without official splitting, we follow the settings in \cite{bi2020rgcnn, deng2022evvgcnn} to randomly select 20$\%$ of data for testing, and the rest is used for training. 

\subsubsection{Implementation Details}

\paragraph{Representation} For N-Caltech101, N-Cars, and ASL-DVS, the entire event stream is converted into the event voxel set representation (Section \ref{Sec3_A_Representation}) as input of our model. For CIFAR10-DVS, substreams of 200 ms are randomly cut for representation during training and evaluation. We fix the compensation coefficient $T$ as 4 for all datasets. Considering the different spatiotemporal sizes of the datasets, we set the size $(H_{\rm v}, W_{\rm v}, T_{\rm v})$ of event voxels as follows: (5, 5, 1) for N-Cars; (10, 10, 1) for other three datasets. We set the number $N_{\rm v}$ of input voxels as: 512 for CIFAR10-DVS, N-Cars, and ASL-DVS; 1024 for N-Caltech101. Moreover, the value $D_{\rm f}$ of MLP for voxel feature encoding is fixed as 32 in all experiments.

\paragraph{Network Details} The architecture of the object classification network is detailed in Section \ref{Sec3_D_Architecture}. In the event voxel transformer encoder, we set the output dimensions $D_{\rm out}$ to 64, 64, and 128 for three MNEL layers, respectively. The total number $N_{\rm n}$ of the nearest neighbors is 24 when aggregating local features, and the downsampling rate $U$ of voxel sampling operations is 0.75. For two VSAL layers, we set the value of $D$ to 128. In the classification head, the output dimensions of three FC layers are 512, 256, and the number of classes. During training, we add two dropout layers with a probability of 0.5 after the first two FC layers to avoid overfitting.

\paragraph{Training Settings} The proposed EVSTr model is implemented using PyTorch. We train the object classification network from scratch for 250 epochs by optimizing cross-entropy loss using the SGD optimizer with a momentum of 0.9. The batch size is 32 for all datasets. We adjust the learning rate from 3e-2 to 1e-6 via cosine annealing.

\begin{table}[!t]
\centering
\renewcommand\arraystretch{1.2}
\setlength{\tabcolsep}{2.7pt}
\caption{Accuracy of Models for Object Classification}
\label{Table_Results_on_Object}
\begin{threeparttable}
\begin{tabular}{lccccc}
\hline \hline
\textbf{Method} & \multicolumn{1}{l}{\textbf{Type$^{\dagger}$}} & \textbf{N-Cal101} & \textbf{CIF10-DVS} & \textbf{N-Cars} & \textbf{ASL-DVS} \\ \hline
\multicolumn{6}{c}{\textbf{Pre-trained on ImageNet}} \\
EST \cite{gehrig2019est} & F & 0.837 & 0.749 & 0.925 & 0.991 \\
E2VID \cite{rebecq2019e2vid} & F & 0.866 & - & 0.910 & - \\
M-LSTM \cite{cannici2020mlstm} & F & 0.857 & 0.730 & 0.957 & 0.992 \\
MVF-Net \cite{deng2021mvfnet} & F & \textbf{0.871} & \textbf{0.762} & \textbf{0.968} & \textbf{0.996} \\ \hline
\multicolumn{6}{c}{\textbf{Without Pre-training}} \\
EST \cite{gehrig2019est} & F & 0.753 & 0.634 & 0.919 & 0.979  \\
M-LSTM \cite{cannici2020mlstm} & F & 0.738 & 0.631 & 0.927 & 0.980 \\
MVF-Net \cite{deng2021mvfnet} & F & 0.687 & 0.599 & 0.927 & 0.971 \\
AsyNet \cite{messikommer2020asynet} & F & 0.745 & 0.663 & 0.944 & - \\
PointNet++ \cite{qi2017pointnet++} & P & 0.503 & 0.465 & 0.809 & 0.947 \\
RG-CNNs \cite{bi2020rgcnn} & P & 0.657 & 0.540 & 0.914 & 0.901 \\
EvS-S \cite{li2021evs} & P & 0.761 & 0.680 & 0.931 & - \\
EV-VGCNN \cite{deng2022evvgcnn} & P & 0.748 & 0.670 & \textbf{0.953} & 0.983 \\
AEGNN \cite{schaefer2022aegnn} & P & 0.668 & - & 0.945 & - \\
VMV-GCN \cite{xie2022vmvgcn} & P & 0.778 & 0.690 & 0.932 & 0.989 \\
ECSNet \cite{chen2022ecsnet} & P & 0.693 & 0.727 & 0.946 & \textbf{0.997} \\
\textbf{Ours} & P & \textbf{0.797} & \textbf{0.731} & 0.941 & \textbf{0.997} \\ \hline \hline
\end{tabular}
\begin{tablenotes}[para,flushleft]
\item \scriptsize $^{\dagger}$The types ``F'' and ``P'' represent frame-based and point-based methods.\\
\item \scriptsize Note: N-Caltech101 \& CIFAR10-DVS are abbreviated as N-Cal101 \& CIF10-DVS. 
\end{tablenotes}
\end{threeparttable}
\end{table}

\begin{figure}[!t]
\centering
\includegraphics[width=3.45in]{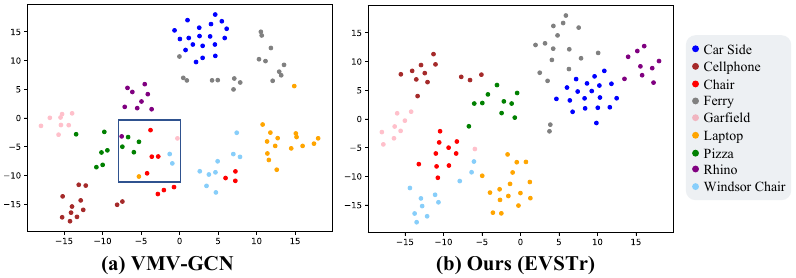}
\caption{The t-SNE visualization of feature representations from VMV-GCN \cite{xie2022vmvgcn} and the proposed EVSTr model on the N-Caltech101 dataset. Best viewed by zooming in.}
\label{Fig_tSNE}
\end{figure}

\subsubsection{Results} As listed in Table \ref{Table_Results_on_Object}, we report the accuracy of our model and state-of-the-art methods on four event-based object classification datasets. To conduct a comprehensive analysis, both point-based \cite{qi2017pointnet++, li2021evs, deng2022evvgcnn, schaefer2022aegnn, xie2022vmvgcn, chen2022ecsnet} and frame-based \cite{gehrig2019est, rebecq2019e2vid, cannici2020mlstm, deng2021mvfnet, messikommer2020asynet} methods are our comparison models. For these frame-based methods (i.e., EST \cite{gehrig2019est}, M-LSTM \cite{cannici2020mlstm} and MVF-Net \cite{deng2021mvfnet}) using pre-trained weights on ImageNet \cite{deng2009imagenet}, we also report their performance by training from scratch for a fair comparison, since the point-based family cannot be pre-trained on large-scale image datasets.

Compared to point-based counterparts, our model outperforms state-of-the-art methods and gains a notable improvement (1.9$\%$ increase) than the second place on the challenging N-Caltech101 dataset, demonstrating the effectiveness of EVSTr. As shown in Fig. \ref{Fig_tSNE}, we further provide the visualization of feature representations learned by VMV-GCN \cite{xie2022vmvgcn} and ours on the testing samples of N-Caltech101 using t-SNE \cite{van2008tsne}. The visualization intuitively shows that our model learns more discriminative spatiotemporal representations because several confusing samples are not distinguished well using the previous state-of-the-art method VMV-GCN, such as the region highlighted in a bounding box. Both experimental results and feature visualization prove the superior representation capability of EVSTr, and we attribute the improvement to two strategies in our model. (\textit{i}) MNEL attentively embeds multi-scale neighbor information into a local representation for each event voxel. The multi-scale attentive aggregation fully explores the positional and semantic relations between neighboring voxels and thus can extract discriminative features. (\textit{ii}) The VSAL layers exploit the long-range dependencies between voxels via feature interaction, allowing us to learn a better global representation than other methods.

Compared to dense frame-based methods using pre-trained models, our method still obtains a competitive performance on ASL-DVS, N-Cars, and CIFAR10-DVS without utilizing prior knowledge from the image domain. When training them from scratch for fair competition, EVSTr achieves higher accuracy than them on all datasets. Besides, our model outperforms the sparse frame-based method AsyNet \cite{messikommer2020asynet} while maintaining lower model complexity. A detailed comparison of model complexity is presented in the following part.

\renewcommand\arraystretch{1.2}
\setlength{\tabcolsep}{4.6pt}
\begin{table}[!t]
\centering
\caption{Model Complexity of Different Methods on Object Classification}
\label{Table_Complexity_on_ObjCls}
\begin{threeparttable}
\begin{tabular}{lcccc}
\hline \hline
\textbf{Method} & \textbf{Type} & \textbf{Params$^{\dagger}$}(M) & \textbf{MACs$^{\dagger}$}(G) & \textbf{Time$^{\ddagger}$}(ms) \\ \hline
EST \cite{gehrig2019est} & F & 21.38 & 4.28 & 6.41 \\
M-LSTM \cite{cannici2020mlstm} & F & 21.43 & 4.82 & 10.89 \\
MVF-Net \cite{deng2021mvfnet} & F & 33.62 & 5.62 & 10.09 \\ 
AsyNet \cite{messikommer2020asynet} & F & 3.70 & - & - \\ \hline
PointNet++ \cite{qi2017pointnet++} & P & 1.76 & 4.03 & 103.85 \\
RG-CNNs \cite{bi2020rgcnn} & P & 19.46 & 0.79 & - \\
EV-VGCNN \cite{deng2022evvgcnn} & P & 0.84 & 0.70 & 7.12 \\
AEGNN \cite{schaefer2022aegnn} & P & 20.40 & - & - \\
VMV-GCN \cite{xie2022vmvgcn} & P & 0.86 & 1.30 & 6.27 \\
\textbf{Ours} & P & 0.93 & 0.34 & 6.62 \\ \hline \hline
\end{tabular}
\begin{tablenotes}[para,flushleft]
\item \scriptsize $^{\dagger}$We report the number of parameters and MACs on N-Caltech101. $^{\ddagger}$The average inference time is measured on N-Cars.
\end{tablenotes}
\end{threeparttable}
\end{table}

\subsubsection{Complexity Analysis} Table \ref{Table_Complexity_on_ObjCls} lists the model complexity of different methods on object classification. We evaluate the model complexity comprehensively by three metrics: the number of trainable parameters, the number of multiply–accumulate operations (MACs), and average inference time. Specifically, we report the number of parameters and MACs on N-Caltech101. Our model achieves the best accuracy on N-Caltech101 while keeping the lowest model complexity (only 0.34G MACs), demonstrating the high efficiency of EVSTr in event-based representation learning. We further measure the average inference time of EVSTr on N-Cars using a workstation (CPU: Intel Core i7, GPU: NVIDIA RTX 3090, RAM: 64GB). Our method needs 6.62 ms to recognize a sample equivalent to a throughput of 151 samples per second, showing the practical potential in high-speed scenarios.

\subsection{Action Recognition} \label{Sec4_B_Action}

\subsubsection{Dataset} UCF101-DVS \cite{bi2020rgcnn} and HMDB51-DVS \cite{bi2020rgcnn} are two converted datasets that record RGB videos by a still event camera. Although these two provide challenging benchmarks for event-based action recognition, they are limited by the frame rate of the source videos and cannot reflect the real-world application scenarios of event cameras. Existing real-world datasets, such as DVS128 Gesture (DvsGesture) \cite{amir2017dvsgesture} and DailyAction-DVS (DailyAction) \cite{liu2021motionsnn}, are collected in relatively simple visual conditions. They are recorded by still cameras in sufficient illumination, lacking dynamic backgrounds and low-light challenges. We address the limitations by presenting the Neuromorphic Human Action Recognition dataset (NeuroHAR) recorded under more challenging conditions, such as low-light illumination and camera motion. The statistics of existing real-world datasets and our proposed NeuroHAR are listed in Table \ref{Table_Action_Dataset}.

\renewcommand\arraystretch{1.2}
\setlength{\tabcolsep}{2pt}
\begin{table}[!t]
\centering
\caption{Statistics of Event-Based Action Recognition Datasets Recorded in the Real-World Environment}
\label{Table_Action_Dataset}
\begin{threeparttable}
\begin{tabular}{lccccccc}
\hline \hline
\textbf{Dataset} & \textbf{Samples} & \textbf{Classes} & \textbf{Duration}(s) & \textbf{Modality$^{*}$} & \textbf{CM$^{\dagger}$} & \textbf{LLI$^{\ddagger}$} \\ \hline
DvsGesture \cite{amir2017dvsgesture} & 1342 & 11 & $\approx$ 6.0 & E & $\times$ & $\times$ \\
DailyAction \cite{liu2021motionsnn} & 1440 & 12 & 1.1 $\sim$ 5.8 & E & $\times$ & $\times$ \\
\textbf{NeuroHAR} & 1584 & 18 & 0.9 $\sim$ 5.7 & E, R, D & $\surd$ & $\surd$ \\ \hline \hline
\end{tabular}
\begin{tablenotes}[para,flushleft]
\item \scriptsize $^{*}$E, R, and D denote event, RGB, and depth modalities. $^{\dagger}$CM denotes camera motion. $^{\ddagger}$LLI denotes low-light illumination.
\end{tablenotes}
\end{threeparttable}
\end{table}

\begin{figure*}[htbp]
\centering
\includegraphics[width=7.1in]{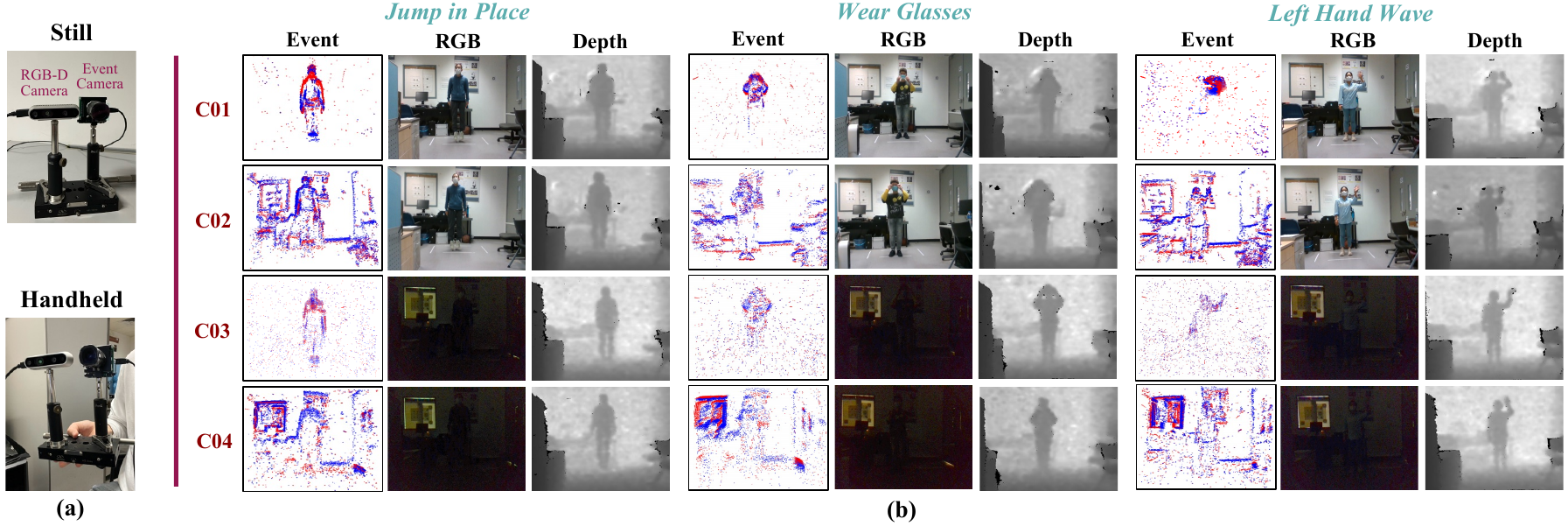}
\caption{(a) Recording system. The system consists of an event camera and an RGB-D camera. (b) Visualization of representative samples on the proposed NeuroHAR dataset. Rows C01 to C04 represent 4 different recording conditions. C01: normal light illumination and still cameras. C02: normal light illumination and moving cameras. C03: low-light illumination and still cameras. C04: low-light illumination and moving cameras. Best viewed by zooming in.}
\label{Fig_Dataset_Visualization}
\end{figure*}

\paragraph{NeuroHAR Dataset Collection} The recording system consists of an event camera (iniLabs DAVIS 240C \cite{brandli2014davis240c}) and an RGB-D camera (Intel RealSense D435), as shown in Fig. \ref{Fig_Dataset_Visualization} (a). The dynamic range of the event camera is 120 dB. We set the frame rate of the RGB-D camera to 30 FPS. We synchronize the time of two vision sensors to align the timestamps of event streams and RGB \& depth video frames. The spatial resolutions of the event and RGB-D cameras are $180 \times 240$ and $480 \times 640$, respectively. The NeuroHAR dataset contains 1584 samples with 3 modalities covering 18 daily human actions. We collect it from 22 subjects under 2 camera states and 2 illumination conditions. The duration of each recording lasts from 0.9 to 5.7 seconds.

Some representative samples of NeuroHAR are visualized in Fig. \ref{Fig_Dataset_Visualization} (b). As a supplement to recording with still cameras, we preserve the background information in the motion scene by hand-held mobile recording (50\% of data per category), making the dataset more suitable for practical applications. Besides, we record human actions in normal light and low-light illumination conditions (50\% low-light data per category) to take advantage of the event camera's high dynamic range. These challenging recording conditions allow NeuroHAR to provide a convincing model evaluation. Each subject performs 18 daily actions in 4 recording conditions, and the categories of NeuroHAR are: \textit{stand}, \textit{sit down}, \textit{walk}, \textit{run}, \textit{jump forward}, \textit{pick up}, \textit{wear glasses}, \textit{take off glasses}, \textit{hand clap}, \textit{throw}, \textit{left hand wave}, \textit{right hand wave}, \textit{left kick}, \textit{right kick}, \textit{jumping jacks}, \textit{answer the call}, \textit{jump in place}, and \textit{carry schoolbag}. These categories contain body movements, human-object interaction (e.g., \textit{carry schoolbag}) and potentially abnormal behavior (e.g., \textit{right kick}), gaining more practical significance. In addition, NeuroHAR provides the recordings with 3 visual modalities, including event, RGB, and depth, which can be also used for multi-modal action recognition with events and video frames. To evaluate the recognition performance on NeuroHAR, the splitting strategy of the training and testing sets is: 16 subjects are designated as the training set, and the remaining 6 subjects are reserved for testing. The NeuroHAR dataset is publicly available\footnote{\href{https://github.com/bochenxie/NeuroHAR}{https://github.com/bochenxie/NeuroHAR}}.

\paragraph{Pre-Processing and Dataset Settings} We evaluate the proposed method on five action recognition datasets, including UCF101-DVS, HMDB51-DVS, DvsGesture, DailyAction, and newly presented NeuroHAR (only using the event modality). For UCF101-DVS and HMDB51-DVS, substreams of 2000 ms are randomly cut for training and evaluation. We adopt the pre-processing method in \cite{wang2019eventcloud, xie2022vmvgcn} to cut 800 ms clips of DvsGesture as input. For DailyAction and NeuroHAR, the entire event stream is fed into our model. We train EVSTr separately on each training set and evaluate it on the testing sets. For UCF101-DVS, HMDB51-DVS, and DailyAction without a training/testing splitting, we follow the settings in \cite{bi2020rgcnn} to randomly select 20$\%$ of data for testing, and the remaining samples are used for training.

\subsubsection{Implementation Details}

\paragraph{Representation} As detailed in Section \ref{Sec3_C_S2TM}, an effective segment modeling strategy is introduced to fit the long-range temporal structure of event streams for action recognition. We spilt an entire event stream into multiple segments for voxel-wise representations individually: $K = 4$ for UCF101-DVS, HMDB51-DVS, DvsGesture, and DailyAction; $K = 6$ for more challenging NeuroHAR. For each stream segment, the compensation coefficient $T$ is fixed as 8 in all datasets. When generating the event voxel set per segment, we set the size of event voxels as follows: (5, 5, 1) for DvsGesture and DailyAction; (10, 10, 1) for others. And we set the number $N_{\rm v}$ of input voxels for each segment as: 512 for DvsGesture; 1024 for other four datasets. The dimension $D_{\rm f}$ of voxel feature encoding is also 32.

\paragraph{Network Details} The architecture of the action recognition network is illustrated in Fig. \ref{Fig_Model}. The model applied to action recognition and object classification keeps the same settings of event voxel transformer encoder, which are detailed in Section \ref{Sec4_A_Object}. We utilize the proposed encoder to extract per-segment features and aggregate them into a final representation by S$^{2}$TM for recognition. In the S$^{2}$TM module, the output dimension of mapping function is 512. A shallow transformer encoder layer \cite{dosovitskiy2021vit} is used for temporal modeling with the following settings: 1-layer depth, 8 heads, 64 for the size of linear transformations in each head, and 1024 for the dimension of the hidden layer in the feed-forward network. The classification head is a single FC layer for predicting action categories.

\paragraph{Training Settings} We train the action recognition network for 300 epochs by optimizing the cross-entropy loss using the SGD optimizer with the batch size of 16. Also, the cosine annealing strategy is used to adjust the learning rate from 1e-2 to 1e-7.

\begin{table*}[!t]
\centering
\renewcommand\arraystretch{1.2}
\setlength{\tabcolsep}{12pt}
\caption{Accuracy of Models for Action Recognition}
\label{Table_Results_on_Action}
\begin{threeparttable}
\begin{tabular}{lcccccc}
\hline \hline
\multirow{2.5}{*}{\textbf{Method}} & \multirow{2.5}{*}{\textbf{Type}} & \multicolumn{2}{c}{\textbf{Converted Datasets}} & \multicolumn{3}{c}{\textbf{Real-World Datasets}}  \\ \cmidrule(lr){3-4} \cmidrule(lr){5-7} 
&  & \textbf{UCF101-DVS} & \textbf{HMDB51-DVS} & \textbf{DvsGesture} & \multicolumn{1}{c}{\textbf{DailyAction}} & \multicolumn{1}{c}{\textbf{NeuroHAR}} \\ \hline
\multicolumn{7}{c}{\textbf{Pre-trained on Kinetics-400}} \\
I3D$^{\dagger}$ \cite{carreira2017i3d} & F & \textbf{0.781} & 0.653 & \textbf{0.983} & 0.962 & 0.854 \\
TANet$^{\dagger}$ \cite{liu2021tam} & F & 0.776 & \textbf{0.665} & 0.973 & \textbf{0.965} & \textbf{0.866} \\
TimeSformer \cite{bertasius2021timesformer} & F & 0.772 & 0.664 & 0.967 & 0.950 & 0.828 \\\hline
\multicolumn{7}{c}{\textbf{Without Pre-training}} \\
I3D$^{\dagger}$ \cite{carreira2017i3d} & F & 0.603 & 0.397 & 0.951 & 0.909 & 0.738 \\
TANet$^{\dagger}$ \cite{liu2021tam} & F & 0.669 & \textbf{0.642} & 0.974 & 0.963 & 0.857 \\
TimeSformer \cite{bertasius2021timesformer} & F & 0.541 & 0.384 & 0.917 & 0.906 & 0.613 \\
RG-CNN + Incep. 3D \cite{bi2020rgcnn} & P & 0.678 & 0.515  & 0.968 & - & - \\
MotionSNN \cite{liu2021motionsnn} & P & - & - & 0.927 & 0.903 & - \\
VMV-GCN \cite{xie2022vmvgcn} & P & - & - & 0.975 & 0.941 & 0.803 \\
ECSNet \cite{chen2022ecsnet} & P & 0.702 & - & \textbf{0.986} & - & - \\
\textbf{Ours} & P & \textbf{0.735} & 0.607 & \textbf{0.986} & \textbf{0.996}  & \textbf{0.894}\\ \hline \hline
\end{tabular}
\begin{tablenotes}[para,flushleft]
\item \scriptsize $^{\dagger}$The model uses ResNet-50 as the backbone.
\end{tablenotes}
\end{threeparttable}
\end{table*}

\subsubsection{Results} We compare our model with state-of-the-art point-based and frame-based methods on event-based action recognition. Motion-based SNN (MotionSNN) \cite{liu2021motionsnn} and graph-based VMV-GCN are selected as representative point-based methods. The experimental results are evaluated with their default settings. Moreover, we re-implement several powerful video classification models on event-based datasets as frame-based counterparts, such as 3D CNN-based I3D \cite{carreira2017i3d}, 2D CNN-based TANet \cite{liu2021tam}, and transformer-based TimeSformer \cite{bertasius2021timesformer}. For I3D and TANet, we use ResNet-50 \cite{he2016resnet} as the backbone. Following the pre-processing in \cite{bi2020rgcnn}, we construct an independent 3-channel frame by \cite{zhu2018voxelgrid} to integrate events within 1/30 s and densely sample 16 event frames as input of frame-based models. We present their results with and without Kinetics-400 \cite{carreira2017i3d} pre-training on these benchmark datasets.

\begin{figure}[!t]
\centering
\includegraphics[width=3.2in]{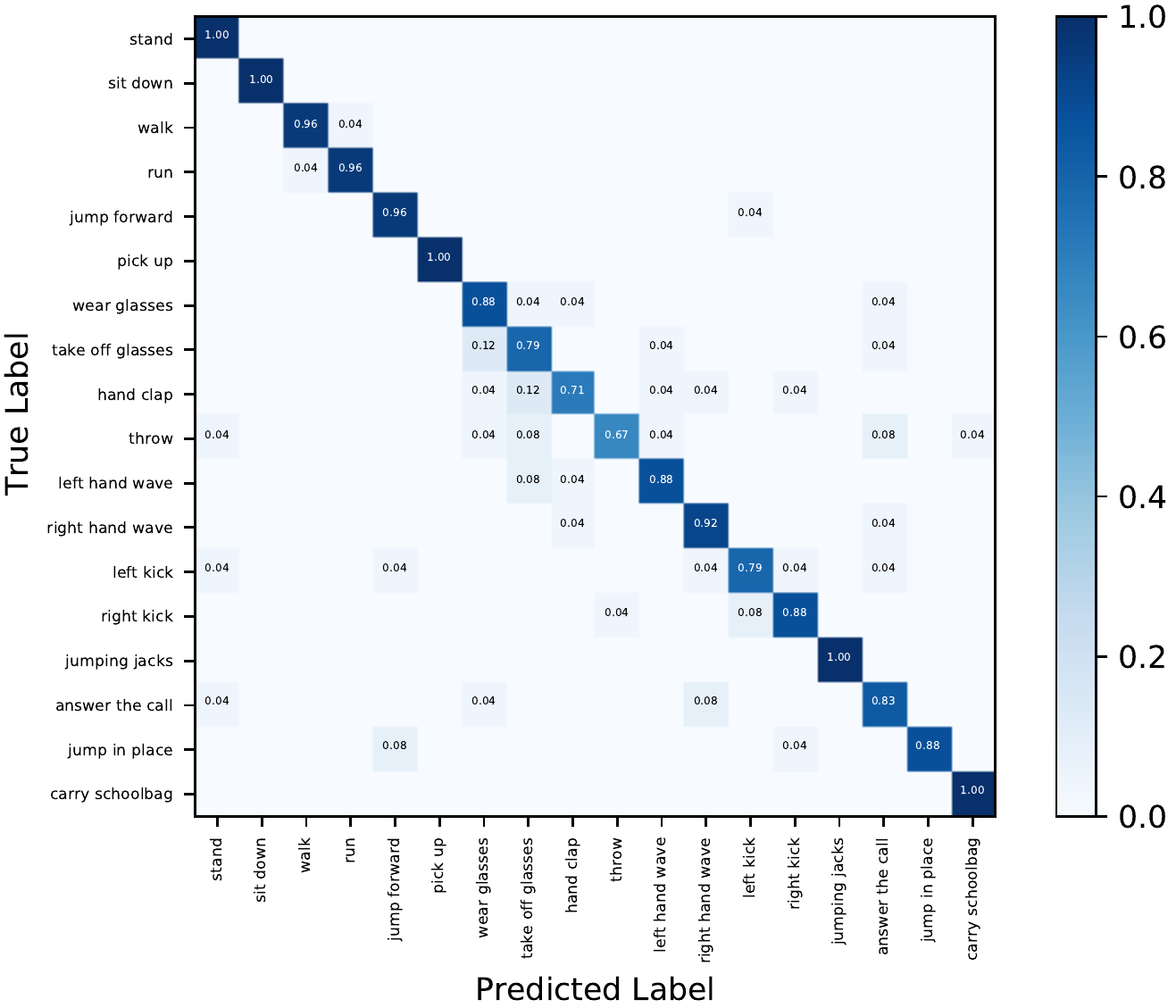}
\caption{Confusion matrix of the proposed EVSTr model on the testing set of NeuroHAR. Best viewed by zooming in.}
\label{Fig_Confusion_Matrix}
\end{figure}

We report the recognition performance of EVSTr and state-of-the-art methods in Table \ref{Table_Results_on_Action}. For the converted datasets UCF101-DVS and HMDB51-DVS, our method achieves the best accuracy among point-based methods and has competitive performance compared to heavyweight frame-based models. On three real-world datasets, the results show that our proposed method obtains the best performance on all testing sets and outperforms the state-of-the-art methods by a large margin. Particularly, EVSTr performs better than these frame-based methods pre-trained on the large-scale video dataset Kinetics-400, such as 3.1$\%$ improvement on DailyAction and 2.8$\%$ improvement on NeuroHAR compared to the second place TANet. This indicates that our proposed method can effectively learn temporal dynamics and motion patterns from long-duration event streams. The notable improvement can be attributed to the effective design of the model. (\textit{i}) The event voxel transformer encoder hierarchically exploits the spatiotemporal relationship between voxels to extract short-term temporal dynamics within a stream segment. (\textit{ii}) The introduced segment modeling strategy endows EVSTr with long-range temporal learning capability, thus significantly improving the recognition accuracy. Furthermore, the confusion matrix of our model on the newly proposed NeuroHAR dataset is shown in Fig. \ref{Fig_Confusion_Matrix}, where our EVSTr model distinguishes most categories clearly. Even so, there is still room for improvement in accuracy, suggesting NeuroHAR as a challenging dataset.

\renewcommand\arraystretch{1.2}
\setlength{\tabcolsep}{2.3pt}
\begin{table}[!t]
\centering
\caption{Model Complexity of Different Models for Action Recognition on DailyAction}
\label{Table_Complexity_on_Action}
\begin{threeparttable}
\begin{tabular}{lcccc}
\hline \hline
\textbf{Method} & \textbf{Type} & \textbf{Params}(M) & \textbf{MACs}(G) & \textbf{Throughput$^{\dagger}$(samp./s)} \\ \hline
I3D \cite{carreira2017i3d} & F & 49.19 & 59.28 & 34\\
TANet \cite{liu2021tam} & F & 24.80 & 65.94 & 32 \\
TimeSformer \cite{bertasius2021timesformer} & F & 121.27 & 379.70 & 23 \\ \hline
VMV-GCN \cite{xie2022vmvgcn} & P & 0.84 & 0.33 & 99 \\
\textbf{Ours} & P & 2.88 & 1.38 & 41 \\ \hline \hline
\end{tabular}
\begin{tablenotes}[para,flushleft]
\item \scriptsize $^{\dagger}$The unit of throughput is the number of processed samples per second.
\end{tablenotes}
\end{threeparttable}
\end{table}

\subsubsection{Complexity Analysis} The model complexity of different action recognition methods on the DailyAction dataset is summarized in Table \ref{Table_Complexity_on_Action}, which includes the number of trainable parameters, the number of MACs, and the inference throughput measured at a batch size of 1. VMV-GCN has the lowest computational complexity and maximum throughput among all methods by processing short-duration clips from event samples, but its recognition performance is much lower than ours. Compared to frame-based methods, our model demonstrates leadership in model complexity because the lightweight EVSTr takes sparse voxel-wise representations as input and thus benefits from less redundant computation.

\subsection{Ablation Study} \label{Sec4_C_Ablation}
In this subsection, we conduct ablation studies to verify the effectiveness of core design strategies in our proposed model.

\renewcommand\arraystretch{1.2}
\setlength{\tabcolsep}{4.8pt}
\begin{table}[t]
\caption{Effectiveness of Different Designs in MNEL}
\label{Table_Diff_Variants_in_MNEL}
\centering
\begin{threeparttable}
\begin{tabular}{cccccc}
\hline \hline
\textbf{Variants} & \textbf{Ev. Rep.} & \textbf{MSE}$^{\dagger}$ & \textbf{AFA}$^{\ddagger}$ & \textbf{N-Caltech101} & \textbf{DailyAction} \\ \hline
A & Voxel & $\times$ & $\times$ & 0.765 & 0.986 \\
B & Voxel & $\surd$ & $\times$ & 0.774 & 0.990 \\
C & Voxel & $\surd$ & $\surd$ & \textbf{0.797} & \textbf{0.996} \\ \hline
D & Point & $\surd$ & $\surd$ & 0.628 & 0.953 \\
E & Voxel-B$^{*}$ & $\surd$ & $\surd$ & \textbf{0.798} & \textbf{0.994} \\ \hline \hline
\end{tabular}
\begin{tablenotes}[para,flushleft]
\item \scriptsize $^{\dagger}$MSE denotes multi-scale embedding in the neighborhood space. $^{\ddagger}$AFA denotes attentive feature aggregation. $^{*}$Voxel-B denotes using bilinear interpolation for event integration.
\end{tablenotes}
\end{threeparttable}
\end{table}

\renewcommand\arraystretch{1.2}
\setlength{\tabcolsep}{6.8pt}
\begin{table}[t]
\caption{Comparison of Different Positional Encoding Methods in VSAL}
\label{Table_Diff_PosEnc_in_VSAL}
\centering
\begin{threeparttable}
\begin{tabular}{ccccc}
\hline \hline
\textbf{Variants} & \textbf{APE}$^{\dagger}$ & \textbf{RPB}$^{\ddagger}$ & \textbf{N-Caltech101} & \textbf{DailyAction} \\ \hline
1 & $\times$ & $\times$ & 0.755 & 0.958 \\
2 & $\surd$ & $\times$ & 0.772 & 0.983 \\
3 & $\times$ & $\surd$ & 0.761 & 0.976 \\
4 & $\surd$ & $\surd$ & \textbf{0.797} & \textbf{0.996} \\ \hline \hline
\end{tabular}
\begin{tablenotes}[para,flushleft]
\item \scriptsize $^{\dagger}$APE denotes absolute positional embeddings. $^{\ddagger}$RPB denotes relative position bias.
\end{tablenotes}
\end{threeparttable}
\end{table}

\subsubsection{Feature Aggregation Strategies in MNEL} Ablations of the multi-scale attentive aggregation in MNEL on object classification (N-Caltech101) and action recognition (DailyAction) are reported in Table \ref{Table_Diff_Variants_in_MNEL}. Variants A-C represent different feature aggregation strategies using the event voxel set as input. Variants D and E take different event representations as input, such as the point-wise representation \cite{bi2020rgcnn} and event voxel with bilinear interpolation integration \cite{zhu2018voxelgrid}.

Variant A is a baseline strategy directly using max pooling to aggregate single-scale features in the neighborhood space. Based on this, variant B aggregates neighbor information separately at multiple scales and then sums them together to obtain multi-scale features. It achieves slight improvements on both tasks, demonstrating that the multi-scale approach enhances local feature extraction. Nevertheless, aggregating neighbor features only considering the positional relationship may fail to obtain robust local features. To this end, our proposed MNEL layer (variant C) exploits both positional and semantic relations to attentively aggregate neighbor features instead of max pooling in variant B. The multi-scale attention aggregation strategy improves the recognition accuracy over the baseline, suggesting the importance of fully modeling relations and the superiority of attentive aggregation.

We also analyze the performance of different event representations on recognition tasks. Variant D has a significant drop in accuracy when using point-based representations, indicating that voxel-wise representations preserve local semantics better. Besides, variant E adds bilinear interpolation integration \cite{zhu2018voxelgrid} to voxel-wise representations. The results show that our adopted direct integration \cite{deng2022evvgcnn} and bilinear interpolation integration have similar performance on recognition tasks. For computational efficiency, we use direct integration in representation.

\subsubsection{Positional Encoding in VSAL} Table \ref{Table_Diff_PosEnc_in_VSAL} compares different positional encoding approaches in VSAL on object classification and action recognition. Without any positional embeddings, variant 1 only uses the vanilla self-attention operator that lacks position information, thus limiting its global modeling capability. Introducing the absolute positional embeddings (variant 2) or relative position bias (variant 3) individually improves recognition accuracy slightly on two tasks. Interestingly, after adding the relative position bias, variant 3 achieves a more considerable gain in action recognition than object classification, indicating that relative position may better express the positional relationship between voxels in non-fixed motion patterns. When using absolute-relative positional encoding, variant 4 (our model) achieves the best recognition performance, showing that the proposed positional encoding strategy effectively assists self-attention in computing semantic aﬃnities between voxels.

\subsubsection{Effectiveness of Segment Modeling Strategy on Action Recognition} We conduct an ablation study to verify the effectiveness of segment modeling strategy (S$^{2}$TM) and discuss its performance with different temporal modeling functions, as listed in Table \ref{Table_Diff_Temporal_in_S2TM}. After using S$^{2}$TM to perform segment modeling, the recognition performance of our model improves significantly, demonstrating that this strategy enables EVSTr to incorporate a long-range temporal structure. Then, we compare three mainstream temporal modeling approaches in S$^{2}$TM for experiments, including average pooling, long short-term memory (LSTM), and self-attention. Regardless of which approach is used, EVSTr achieves competitive accuracy compared to state-of-the-art models, reflecting the generalization of our segment modeling strategy. On the relatively simple datasets DvsGesture and DailyAction, the basic average pooling even outperforms advanced LSTM. However, on NeuroHAR with more complex temporal dynamics, the learning-based functions, self-attention and LSTM, will lead to better recognition performance. This fact suggests that: (\textit{i}) Adaptive temporal modeling based on self-attention enables efficient learning of complex temporal dynamics. (\textit{ii}) The proposed NeuroHAR is a challenging dataset that can comprehensively evaluate the performance of event-based models. Therefore, we default to self-attention as the temporal modeling function in S$^{2}$TM.

\renewcommand\arraystretch{1.2}
\setlength{\tabcolsep}{3.2pt}
\begin{table}[t]
\caption{Effectiveness of Segment Modeling Strategy on Action Recognition}
\label{Table_Diff_Temporal_in_S2TM}
\centering
\begin{tabular}{ccccc}
\hline \hline
\textbf{S$^{2}$TM} & \textbf{Temporal Modeling} & \textbf{DvsGesture} & \textbf{DailyAction} & \textbf{NeuroHAR} \\ \hline
$\times$ & - & 0.976 & 0.954 & 0.809 \\ \hline
$\surd$ & Average Pooling & 0.979 & 0.993 & 0.840 \\
$\surd$ & LSTM & 0.976 & 0.983 & 0.884 \\
$\surd$ & Self-attention & \textbf{0.986} & \textbf{0.996} & \textbf{0.894} \\ \hline \hline
\end{tabular}
\end{table}

\renewcommand\arraystretch{1.2}
\setlength{\tabcolsep}{3.2pt}
\begin{table}[t]
\caption{Results of Different Input Modalities on NeuroHAR}
\label{Table_Diff_Modalities}
\centering
\begin{tabular}{lccccc}
\hline \hline
\textbf{Method} & \textbf{Modality} & \textbf{Pre-trained} & \textbf{Accuracy} & \textbf{Params(M)} & \textbf{MACs(G)} \\ \hline
I3D \cite{carreira2017i3d} & RGB & $\surd$ & 0.792 & 46.20 & 59.28 \\
I3D \cite{carreira2017i3d} & Depth & $\surd$ & 0.835 & 46.18 & 55.51 \\ \hline
I3D \cite{carreira2017i3d} & Event & $\surd$ & 0.854 & 46.20 & 59.28 \\
I3D \cite{carreira2017i3d} & Event & $\times$ & 0.738 & 46.20 & 59.28 \\ \hline
\textbf{Ours} & Event & $\times$ & \textbf{0.894} & \textbf{2.89} & \textbf{2.08} \\ \hline \hline
\end{tabular}
\end{table}

\subsubsection{Comparison of Input Modalities on NeuroHAR} The proposed NeuroHAR dataset provides multi-modal recordings, including event, RGB, and depth. To demonstrate the advantages of event cameras, we design an experiment to compare the action recognition results when using different modalities as input, as reported in Table \ref{Table_Diff_Modalities}. We take the widely used video classification model I3D (w/ ResNet-50) as a competitor and feed different modalities into it to observe variations in recognition accuracy. The results indicate that the event modality outperforms other modalities, such as +6.2\% than RGB and +1.9\% than depth. Thanks to the high dynamic range and asynchronous sensing characteristics, event cameras can provide higher quality recordings in challenging visual scenes (e.g., low light and motion blur) than other frame-based cameras. It is worth noting that the competitive results achieved by I3D on NeuroHAR benefit from the powerful transfer learning capabilities of CNNs. Without using Kinetics-400 pre-training, the performance of I3D using the event modality suffers a huge drop (i.e., -11.6\%). Compared to the I3D with dense frame input, our EVSTr achieves the best accuracy with only 6.3\% of the trainable parameters and approximately 3.5\% of the MACs of I3D, and without pre-training weights. This demonstrates that the sparse spatiotemporal learning strategy of EVSTr fits the nature of event data, thus achieving leading accuracy and low computational complexity.
% -------- Experiments: End -------- %

% ------------- Conclusion: begin ------------- %
\section{Conclusion}

This work introduces a novel attention-aware model (EVSTr) for spatiotemporal representation learning on event streams. EVSTr takes event voxel sets as input to fit the sparsity of event data and hierarchically learns robust representations for recognition tasks. The proposed event voxel transformer encoder, consisting of MNEL and VSAL, efficiently extracts spatiotemporal features from event voxels in a local-to-global manner. To model the long-range temporal structure for action recognition, a segment modeling strategy (S$^{2}$TM) is designed to learn temporal dynamics from a sequence of segmented voxel sets. Comprehensive experiments on object classification and action recognition show that EVSTr achieves significant accuracy gains and impressively low model complexity compared to state-of-the-art methods. Moreover, we present a new event-based action recognition dataset (NeuroHAR) with multiple modalities recorded in challenging visual scenarios, allowing convincing evaluation for event-based models. In the future, we plan to extend our model to multi-modal action recognition with events and image frames, unlocking the potential of event-image complementarity.
% ------------- Conclusion: end ------------- %

% ------------- Acknowledgments: begin ------------- %
% \section*{Acknowledgments}
% This should be a simple paragraph before the References to thank those individuals and institutions who have supported your work on this article.
% ------------- Acknowledgments: end ------------- %

% ------------- Reference: begin ------------- %
\bibliographystyle{IEEEtran}
\bibliography{IEEEabrv, reference}
% ------------- Reference: end ------------- %

% ------------- Biography: begin ------------- %
\vspace{-8 mm}

\begin{IEEEbiography}
[{\includegraphics[width=1in,height=1.25in,clip,keepaspectratio]{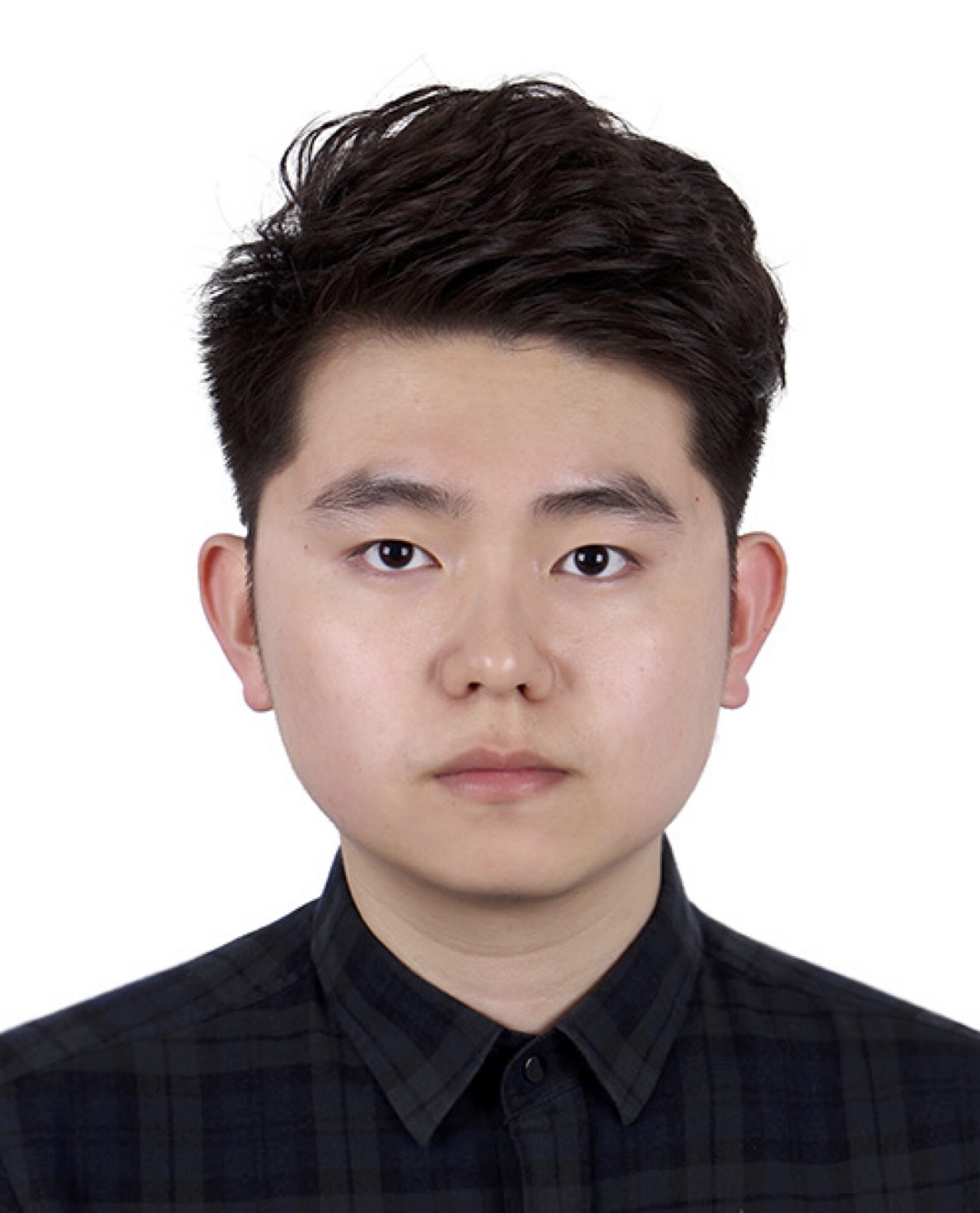}}]{Bochen Xie}
(Member, IEEE) received the B.Eng. degree in automation from the China University of Petroleum (East China), Qingdao, China, in 2019, and the M.Sc. degree in mechanical engineering from the City University of Hong Kong, Hong Kong, in 2020, where he is currently pursuing the Ph.D. degree with the Department of Mechanical Engineering. His research interests include computer vision, pattern recognition, and multi-modal learning.
\end{IEEEbiography}

\vspace{-8 mm}

\begin{IEEEbiography}[{\includegraphics[width=1in,height=1.25in,clip,keepaspectratio]{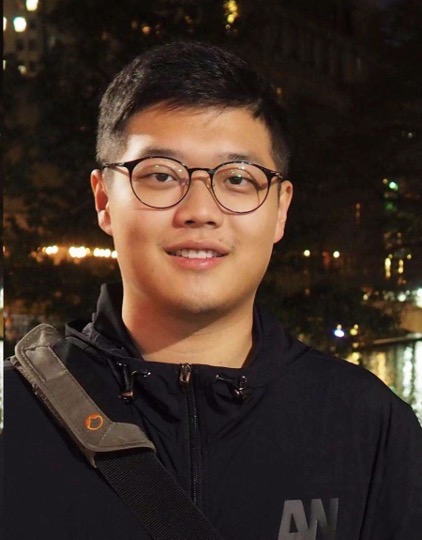}}]{Yongjian Deng}
(Member, IEEE) received the Ph.D. degree from the City University of Hong Kong in 2021. He is currently an Assistant Professor in the College of Computer Science, Beijing University of Technology. His research interests include pattern recognition and machine learning with event cameras.
\end{IEEEbiography}

\vspace{-8 mm}

\begin{IEEEbiography}[{\includegraphics[width=1in,height=1.25in,clip,keepaspectratio]{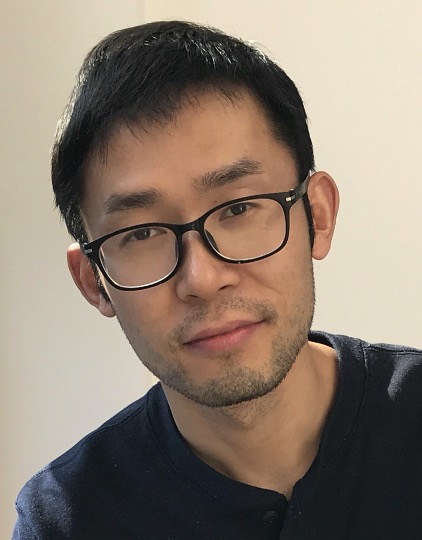}}]{Zhanpeng Shao}
(Member, IEEE) received the B.S. and M.S. degrees in mechanical engineering from the Xi’an University of Technology, Xi’an, China, in 2004 and 2007, respectively, and the Ph.D. degree in computer vision from the City University of Hong Kong, Hong Kong, in 2015.
From 2015 to 2016, he was a Senior Research Associate with the Shenzhen Research Institute, City University of Hong Kong. From 2018 to 2019, he was a Post-Doctoral Research Fellow with the Department of Computing Science, University of Alberta, Edmonton, Canada. From 2016 to 2022, He was an Associate Professor with the College of Computer Science and Technology, Zhejiang University of Technology, China. He is currently an Associate Professor with the College of Information Science and Engineering, Hunan Normal University, China. His current research interests include computer vision, pattern recognition, machine learning, and robot sensing. He received the Best Conference Paper Award on ICMA2014 and ICMA2016.
\end{IEEEbiography}

\vspace{-8 mm}

\begin{IEEEbiography}
[{\includegraphics[width=1in,height=1.25in,clip,keepaspectratio]{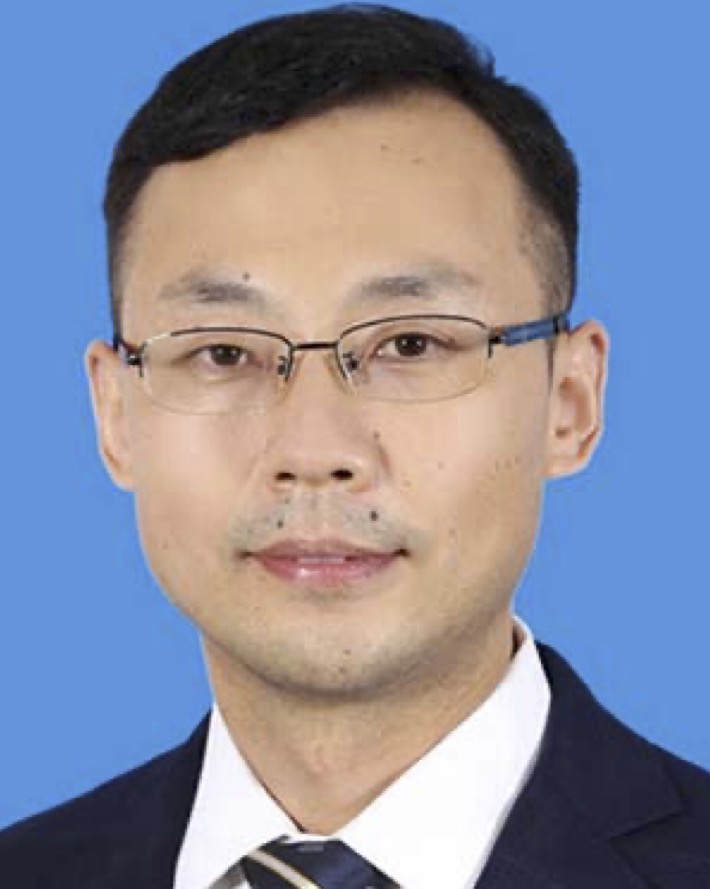}}]{Qingsong Xu}
(Senior Member, IEEE) received the B.S. (Hons.) degree in mechatronics engineering from the Beijing Institute of Technology, Beijing, China, in 2002, and the M.S. and Ph.D. degrees in electromechanical engineering from the University of Macau, Macau, China, in 2004 and 2008, respectively.

He was a Visiting Scholar with the Swiss Federal Institute of Technology, Zurich, Switzerland; the National University of Singapore, Singapore; RMIT University, Melbourne, VIC, Australia; and the University of California at Los Angeles, Los Angeles, CA, USA. He is currently a Full Professor of electromechanical engineering with the University of Macau, where he directs the Smart and Micro/Nano Systems Laboratory. His current research interests include microsystem/nanosystem, micromechatronics/nanomechatronics, robotics and automation, smart materials and structures, and computational intelligence.

Dr. Xu is a Fellow of the American Society of Mechanical Engineers. He is currently an Associate Editor for the {\sc IEEE Transactions on Robotics}, {\sc IEEE Transactions on Automation Science and Engineering}, and \textit{International Journal of Advanced Robotic Systems}, and an Editorial Board Member of the \textit{Chinese Journal of Mechanical Engineering}. He was a Technical Editor for the {\sc IEEE/ASME Transactions on Mechatronics} and an Associate Editor for the {\sc IEEE Robotics and Automation Letters}.
\end{IEEEbiography}

\vspace{-8 mm}

\begin{IEEEbiography}[{\includegraphics[width=1in,height=1.25in,clip,keepaspectratio]{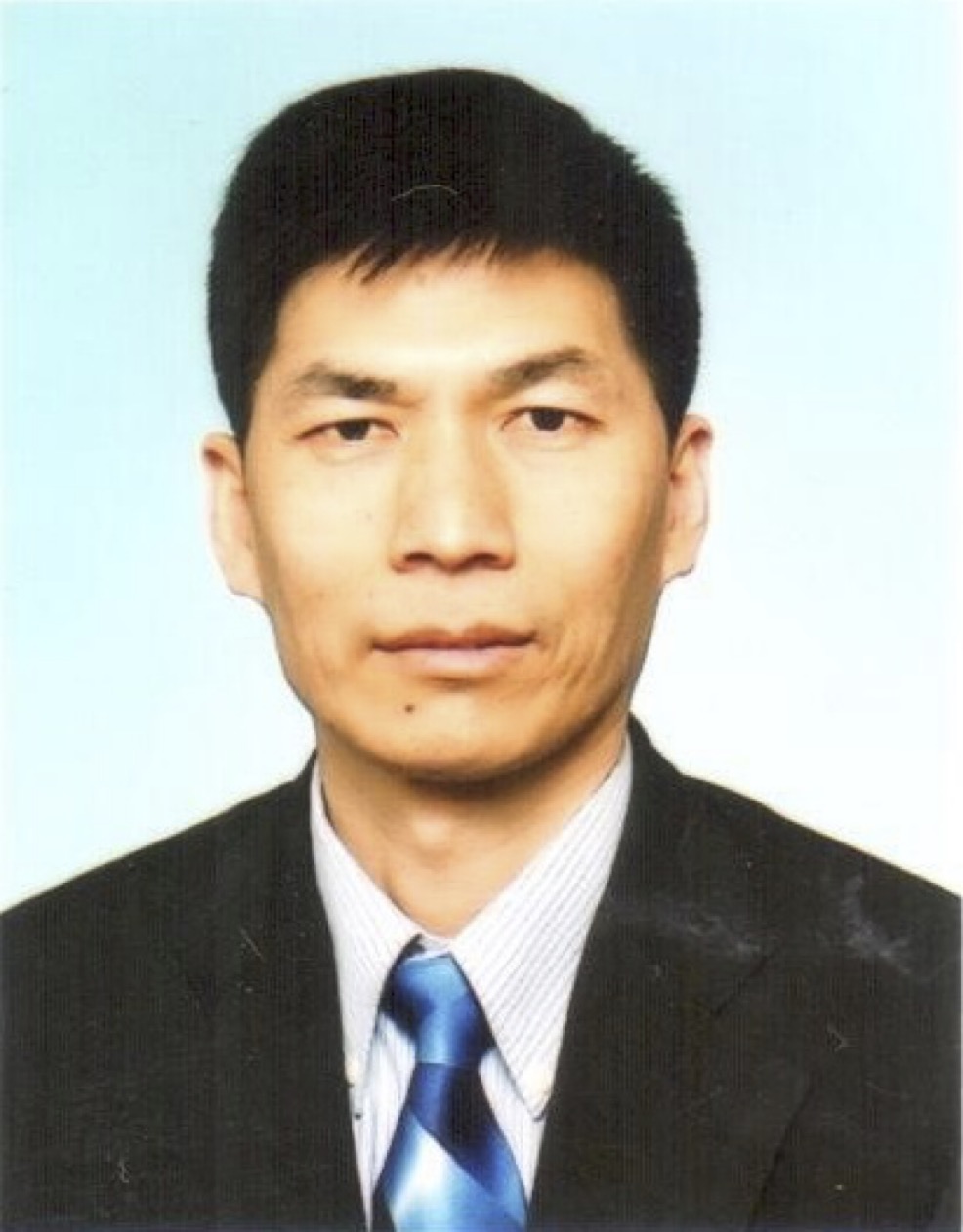}}]{Youfu Li}
(Fellow, IEEE) received the Ph.D. degree in robotics from the Department of Engineering Science, University of Oxford, U.K., in 1993. From 1993 to 1995, he was a Research Staff with the Department of Computer Science, University of Wales, Aberystwyth, U.K. He joined the City University of Hong Kong in 1995, where he is currently a Professor with the Department of Mechanical Engineering. His research interests include robot sensing, robot vision, and visual tracking. In these areas, he has published over 180 articles in refereed international journals. He has served as an Associate Editor for the {\sc IEEE Transactions on Automation Science and Engineering} and \textit{IEEE Robotics and Automation Magazine}, an Editor for the IEEE Robotics and Automation Society's Conference Editorial Board, and a Guest Editor for \textit{IEEE Robotics and Automation Magazine}.
\end{IEEEbiography}
% ------------- Biography: end ------------- %

\end{document}